\documentclass{article} %
\usepackage{iclr2027_conference,times}

\usepackage{amsmath,amsfonts,bm}

\def\eqref#1{equation~\ref{#1}}

\def\1{\bm{1}}

\DeclareMathAlphabet{\mathsfit}{\encodingdefault}{\sfdefault}{m}{sl}
\SetMathAlphabet{\mathsfit}{bold}{\encodingdefault}{\sfdefault}{bx}{n}

\usepackage[table]{xcolor}
\usepackage{multirow}
\usepackage{graphicx}   %
\usepackage{booktabs}  
\usepackage{caption}
\usepackage{subcaption}
\usepackage{hyperref}
\usepackage{url}
\usepackage{amssymb}
\usepackage{enumitem}
\newcommand{\hl}{\cellcolor{cyan!10}}
\usepackage{float}
\usepackage{wrapfig}

\title{X Marks the Spot: \\ Diffusion for Image Geolocalization \\ with Location Representations}

\usepackage{xspace}

\newcommand{\nametext}{DIG}

\newcommand{\name}{\nametext\xspace}

\iclrfinalcopy
\author{%
  Brian Wei \quad
  Chih-Chuan Hsu \quad
  Tianyi Gao \quad
  Daniel Cher \\
  \bfseries Eric Xing \quad
  Danielle Beaulieu \quad
  Nathan Jacobs \\[4pt]
  Washington University in St. Louis \\
}

\begin{document}

\maketitle
\lhead{Preprint}

\begin{abstract}
Image geolocalization predicts the location at which a photo was taken. Accurate geolocalization requires discriminating between both physically and semantically adjacent areas on Earth. However, prior diffusion-based approaches predict locations by generating coordinates, which are semantically unstable over space. We avoid this limitation by learning geolocalization behavior in the pretrained representation spaces of location encoders, which provide both positional \emph{and} semantic understanding. To make this space usable for coordinate prediction, we introduce a \textbf{Location Autoencoder (LAE)}, which pairs a frozen location encoder with a learned decoder mapping embeddings back to coordinates with kilometer-level accuracy worldwide. In the LAE space, we train \textbf{\name}, a latent diffusion model that transports noise to location embeddings conditioned on image features. \name consistently outperforms coordinate-based generative models and sets a new state of the art for parametric geolocalization on three benchmarks: OSV-5M, Im2GPS3k, and YFCC4k. Notably, \name is the first generative model to outperform retrieval-based methods, and further improves retrieve-and-rerank systems when used as a prior. Code and models will be released.
\end{abstract}

\begin{figure}[H]
    \centering
    \includegraphics[width=\textwidth]{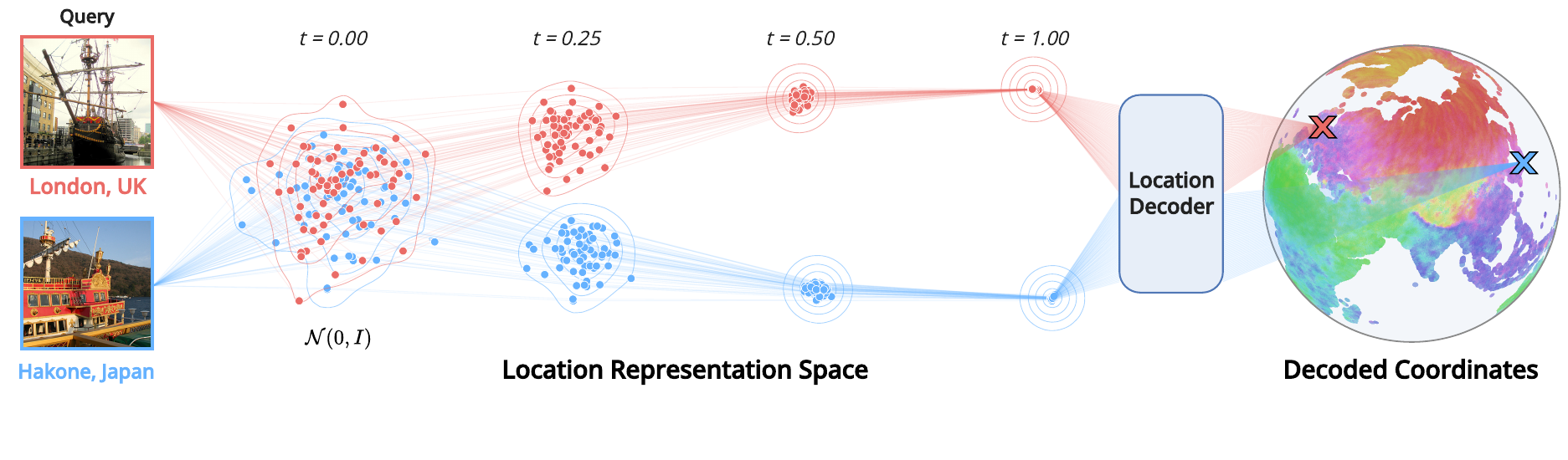}
    \vspace{-30pt}
    \caption{We propose \textbf{D}iffusion for \textbf{I}mage \textbf{G}eolocalization (\textbf{\name}) in the location representation space. Conditioned on the image, a rectified flow transports noise to location embeddings, which a learned decoder maps to coordinates on the globe.}
    \label{fig:teaser}
\end{figure}

\section{Introduction}
\label{sec:intro}

Locating where a photo was taken supports numerous investigative tasks, from forensic applications to human trafficking~\citep{yokota2020revisited, bamigbade2024computer}, media verification~\citep{tahmasebzadeh2023multimodal}, and species range mapping~\citep{mac2019presence, cole2023spatial}.
Yet many photographs are never geotagged~\citep{bamigbade2024computer}, making predictive geolocalization necessary to recover location information from only visual content. Progress in geolocalization performance has advanced dramatically from pre-deep learning~\citep{hays2008im2gps} to large transformer architectures~\citep{seo2018cplanet} and retrieval augmented approaches~\citep{haas2024pigeon}. 

The problem remains challenging because a photograph is often ambiguous. An image in a forest or of a suburban street can be consistent with many places across the globe~\citep{dufour2025around}. Nonetheless, classification- and retrieval-based methods predict a single location from a predefined grid cell~\citep{vo2017revisiting, seo2018cplanet} or a reference gallery~\citep{vivanco2023geoclip, gadi2026hierloc}, respectively. 
Diffusion-based models remove these constraints by learning a probability distribution over continuous coordinates conditioned on the image~\citep{dufour2025around, wang2025locdiff}. 
These generative approaches denoise coordinates on the sphere~\citep{dufour2025around} or a spherical harmonic encoding of the location~\citep{wang2025locdiff}, adopting diffusion spaces that respect the geometry of the Earth. However, since these spaces encode position alone, two locations are similar only if they are nearby. Imagery behaves differently: a pine forest in Canada looks like a pine forest in Siberia, and a city block looks nothing like farmland a few kilometers away. 

Location encoders learn representations with precisely this structure. They map a latitude--longitude pair to a high-dimensional embedding through a positional encoding and a network trained to align with co-located imagery~\citep{vivanco2023geoclip, klemmer2025satclip, cher2026tte}. The positional encoding resolves fine spatial detail, and the alignment maps visually similar environments to nearby embeddings, even when they are far apart on the globe~\citep{klemmer2025satclip, cher2026tte}. In image generation, recent work shows that frozen, pretrained vision foundation models can serve directly as diffusion spaces, achieving state-of-the-art visual quality~\citep{zheng2026rae, tong2026scaling}.
We explore whether the same holds for geolocalization: \emph{can pretrained location representations serve as a diffusion space for generative geolocalization?}

Predicting geolocations by generating location embeddings requires an inverse mapping from representations back to coordinates. To this end, we introduce \emph{Location Autoencoders} (LAEs), which pair a frozen pretrained location encoder with a learned decoder that recovers original locations. 
In the LAE latent space, we train \name, a diffusion model that generates a location embedding conditioned on image features (Figure~\ref{fig:teaser}). 
We evaluate this approach with three location encoders, GeoCLIP, SatCLIP, and TTE. Our contributions are as follows:

\begin{itemize}[leftmargin=*]
\item We propose \textbf{\name}, a latent diffusion model for image geolocalization in the location representation space. Using a simple architecture and denoising objective, we show that changing the generation space from coordinates to location embeddings improves geolocalization accuracy at every scale. 
\item We introduce \textbf{Location Autoencoders (LAEs)}, a pairing of pretrained location encoders with learned coordinate decoders that reconstruct to kilometer-level accuracy across the globe, enabling coordinate prediction. 
\item We show that \name sets a new state of the art for parametric methods on OSV-5M, Im2GPS3k, and YFCC4k, and its samples can further serve as priors to improve retrieve-and-rerank systems.
\end{itemize}

\section{Related Works}
\label{sec:related}

\paragraph{Global Visual Geolocalization.} 
Image geolocalization models can primarily be categorized as classification-, retrieval-, or MLLM-based. Classification methods~\citep{weyand2016planet, seo2018cplanet, muller2018isn, pramanick2022translocator, clark2023geodecoder, haas2024pigeon} partition the Earth into cells and predict one. Retrieval methods~\citep{haas2023streetclip, vivanco2023geoclip, gadi2026hierloc, jia2025georanker} match the query against a geotagged database and return the top candidate's location. MLLM-based methods~\citep{li2024georeasoner, zhou2024img2loc, globe2025, 2026geoagent, SpotAgent2026, li2026reverse, yu2026locatability, jia2026georouter, GeoSearch} reason with multimodal large language models, often with search and map tools. However, classification and retrieval are bounded by the geographic distribution of their cells or reference database, while MLLM-based methods require multi-step inference through large models.

Recently, generative approaches have offered an alternative. Rather than predicting a single point, they capture visual ambiguity by modeling a distribution over locations. \citet{dufour2025around} uses Riemannian flow matching in coordinates on the sphere while LocDiff~\citep{wang2025locdiff} diffuses in multi-scale spherical harmonics. Both trail strong retrieval-based methods, which we attribute to the diffusion spaces being defined by position alone. We instead propose a diffusion model that operates in \textit{semantic} location representation spaces.

\paragraph{Location Representation Learning.}
Location encoders map a coordinate to a high-dimensional representation through a positional encoding~\citep{mac2019presence, space2vec_iclr2020, mai2023sphere2vec, russwurm2024geographic} followed by a neural network. The network is trained to predict downstream targets~\citep{mac2019presence, cole2023spatial, cher2026chronosphere} or contrastively aligned with co-located ground-level~\citep{vivanco2023geoclip} or overhead imagery~\citep{dhakal2024sat2cap, klemmer2025satclip, cher2026tte, dhakal2025range, liu2026gair}. These learned representations are shown to be semantically richer than raw coordinates or the positional encoding alone, and outperform both as features for downstream prediction~\citep{klemmer2025satclip, dhakal2025range}. In these works, the embedding is an input, either a feature for a predictor or a
condition for a generative model~\citep{sastry2026terradit, wei2026terraditomega}. We instead treat it as an output space: we invert the encoder to recover coordinates and generate directly in its embedding space.

\paragraph{Representations for Generation.} Image generation models have progressed remarkably to achieve high visual fidelity~\citep{rombach2022high, peebles2023dit, ma2024sit, baade2026latent}. Recent works~\citep{yu2024repa, singh2025irepa, shi2026svg, yao2025va-vae} align diffusion features with pretrained vision representations during training for improved generation. RAE~\citep{zheng2026rae, singh2026raev2} goes further, diffusing directly in the vision representation space. Subsequent works adopt this paradigm and achieve improved performance on novel-view synthesis~\citep{jang2026repurposingmv} and video generation~\citep{guo2026vrae}. We extend this generative framework by diffusing in pretrained location representations for image geolocalization.

\section{Recovering Coordinates from Location Representations}
\label{sec:lae}
Diffusion for image geolocalization in a location representation space requires a means of recovering coordinates from the generated embeddings.
Inspired by RAE~\citep{zheng2026rae}, we construct a \textbf{Location Autoencoder (LAE)} that pairs a frozen location encoder $E$ with a learnable coordinate decoder $D_\psi$ that maps embeddings back to coordinates.

\begin{figure}[t]
\centering
\begin{minipage}[t]{0.46\textwidth}
    \centering
    \vspace{0pt}
    \captionsetup{type=figure}
    \includegraphics[width=\linewidth]{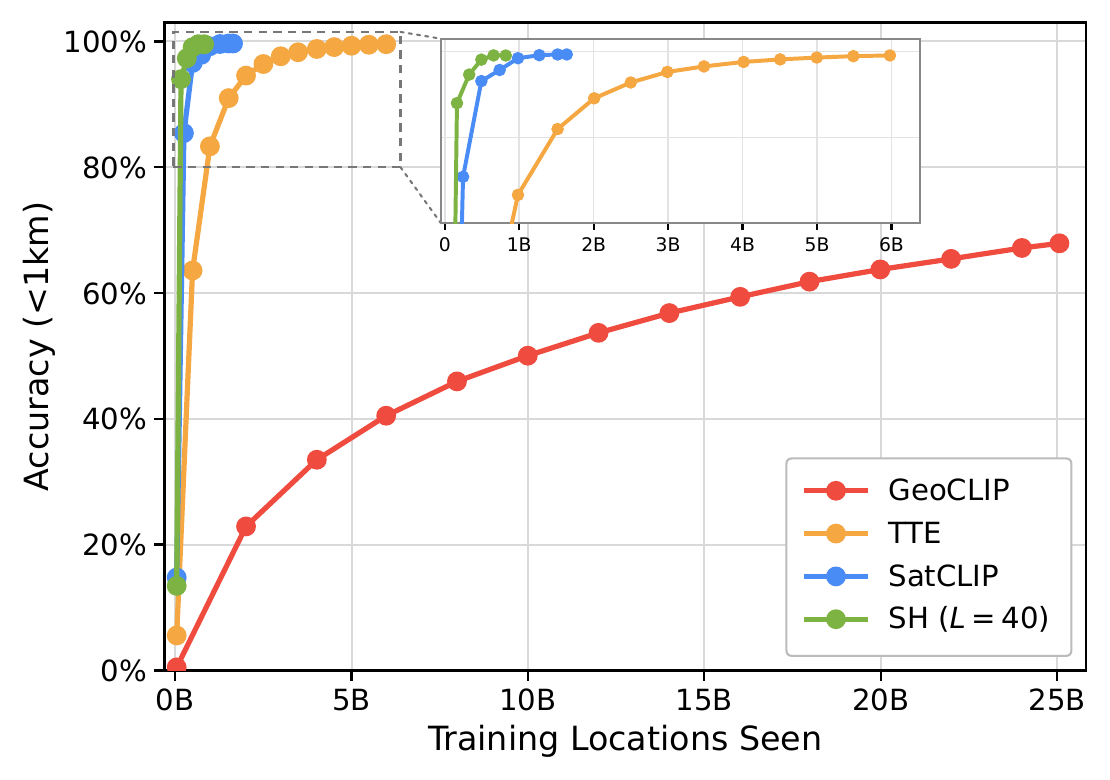}
    \vspace{-21pt}
    \caption{\textbf{Decoding different encoders.}}
    \label{fig:dec_scale}
\end{minipage}%
\hfill
\begin{minipage}[t]{0.50\textwidth}
    \centering
    \vspace{0pt}
    \captionsetup{type=table}
    \scriptsize
    \caption{\textbf{Coordinate reconstruction.} Fraction of 1M land coordinates localized within distances. (a) Choice of location encoder. (b) Analytic refinement on GeoCLIP decoder.}
    \label{tab:lae_ablations}
    \vspace{-3pt}
    \setlength{\tabcolsep}{2.5pt}
    \begin{minipage}[t]{0.42\linewidth}
        \centering
        \vspace{0pt}
        \begin{tabular}{@{}lcc@{}}
            \toprule
            Encoder & $<$1\,km & $<$25\,km \\
            \midrule
            SH ($L=10$) & 99.80 & 100.00 \\
            SH ($L=20$) & 99.73 & 100.00 \\
            SH ($L=40$) & 99.74 & 100.00 \\
            \midrule
            \rowcolor{gray!15} GeoCLIP & 67.91 & 100.00 \\
            SatCLIP & 99.66 & 100.00 \\
            TTE     & 99.60 & 100.00 \\
            \bottomrule
        \end{tabular}
        \vspace{1pt}
        \subcaption{Encoder choice.}
        \label{tab:lae_encoder}
    \end{minipage}%
    \hfill
    \begin{minipage}[t]{0.48\linewidth}
        \centering
        \vspace{0pt}
        \begin{tabular}{@{}lcc@{}}
            \toprule
            Decoder & $<$1\,km & GFLOPs \\
            \midrule
            \rowcolor{gray!15} GeoCLIP & 67.91 & 0.076 \\
            \midrule
            \; +1 LM  & 99.10 & 0.208 \\
            \; +2 LM   & 99.67 & 0.341 \\
            \; +3 LM & 99.79 & 0.473 \\
            \; +4 LM  & 99.85 & 0.605 \\
             \; +5 LM   & 99.89 & 0.737 \\
            \bottomrule
        \end{tabular}
        \vspace{5pt}
        \subcaption{Analytic refinement.}
        \label{tab:lae_refine}
    \end{minipage}
\end{minipage}
\end{figure}

\paragraph{Inverting location representations.}
Location encoders share a common structure: a fixed positional encoding (e.g., spherical harmonics~\citep{russwurm2024geographic}) that counteracts the spectral bias of neural networks~\citep{rahaman2019spectral,tancik2020fourier}, followed by a learned network. 
The inverse problem, recovering coordinates from an embedding, has received less attention. Because the learned network is non-linear, inversion has been viewed as difficult~\citep{wang2025locdiff}. The output space, however, is compact. Every location lies on the sphere $\mathcal{S}^2$, so densely sampling coordinates can approximate it. We therefore learn a decoder $D_\psi : \mathbb{R}^d \to \mathcal{S}^2$ to recover coordinates from location embeddings via direct regression. We instantiate this network as a simple residual MLP, trained with the objective to minimize geodesic distance between predicted and true locations,
\begin{equation}
  \mathcal{L}_{\text{dist}} = \mathbb{E}_{\ell}\,\big[
  d_{\mathcal{S}^2}\big( D_\psi(x),\, \ell \big)\big],
\end{equation}
where $\ell \in \mathcal{S}^2$ is a location sampled uniformly over land, $x=E(\ell)$ is the location embedding, and $d_{\mathcal{S}^2}$ is the great-circle distance in kilometers, as seen in Figure~\ref{fig:method}. Full architecture and training details are in App.~\ref{app:lae}. In Figure~\ref{fig:dec_scale} and Table~\ref{tab:lae_ablations}, we evaluate the coordinate reconstruction accuracy within distance thresholds ($<1$km, $<25$km) over 1 million randomly held out land coordinates. 

\paragraph{Decoding difficulty depends on the encoder.}
Figure~\ref{fig:dec_scale} and Table~\ref{tab:lae_encoder} compare decoders trained to recover coordinates from spherical harmonics (SH) and three learned location encoders (GeoCLIP, SatCLIP, TTE). All decoders recover coarse position easily (100\% at 25\,km) but differ sharply at the kilometer scale. SH and SatCLIP exceed 99\% after 0.5B and 0.9B training locations, TTE after 4.3B, whereas GeoCLIP is at 67.91\% after 25B with diminishing returns (${<}1$ point per additional 2B). We attribute this to the spatial frequency of the encoding. SH is band-limited, SatCLIP is a smooth network on that basis, and TTE interpolates over learned sites, so coordinate inversion is smooth. GeoCLIP's random Fourier features encode far higher frequencies, forcing the decoder to learn a correspondingly fine-grained inverse, which requires orders of magnitude more training.

\paragraph{Analytic refinement.}
Instead of training the GeoCLIP decoder on billions more locations, we exploit that the encoder $E$ is known, cheap, and differentiable, and use it to refine decoder estimates that are already correct at a coarse scale. Starting from $\hat{\ell}_0 = D_\psi({x})$, we minimize the encoder residual $\|{x} - E(\ell)\|^2$ with Levenberg--Marquardt (LM) steps~\citep{levenberg1944method,marquardt1963algorithm},
\begin{equation}
  \hat{\ell}_{k+1} = \hat{\ell}_k \oplus \big(J_k^{\top} J_k + \lambda_k \operatorname{diag}(J_k^{\top} J_k)\big)^{-1} J_k^{\top}\big({x} - E(\hat{\ell}_k)\big),
  \qquad J_k = \frac{\partial E}{\partial \ell}\Big|_{\hat{\ell}_k} \in \mathbb{R}^{d \times 2},
  \label{eq:lm}
\end{equation}
where $J_k$ is computed exactly by forward-mode automatic differentiation and $\oplus$ applies the 2-D step along the sphere. Exact details are in App~\ref{app:lae_implementation}. We find this refinement drastically improves fine-grained reconstruction at a small cost (Table~\ref{tab:lae_refine}). Notably, a single refinement step raises the GeoCLIP decoder's accuracy at 1\,km from 67.91\% to 99.10\%. We use five refinement steps for our GeoCLIP decoder in all subsequent experiments.

\section{Latent Diffusion with Location Autoencoders}
\label{sec:ldm}

\begin{figure}[htbp]
    \centering
    \includegraphics[width=\textwidth]{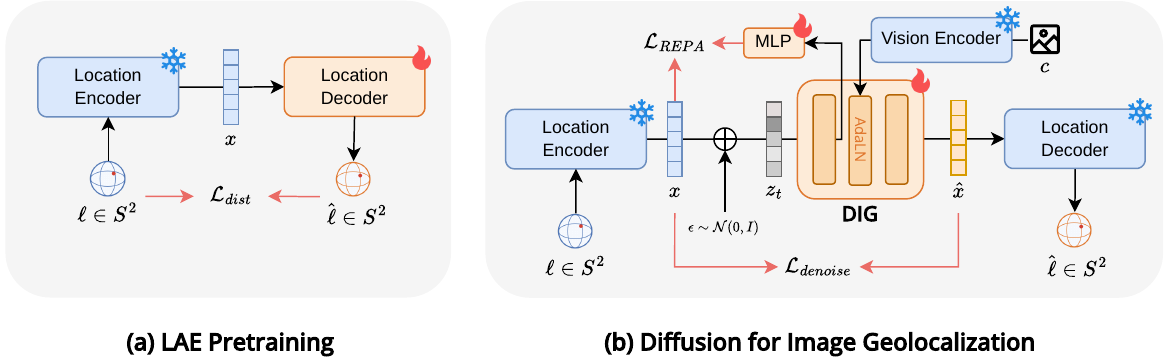}
    \caption{\textbf{Overview of our approach.} (a) Location Autoencoder (LAE) pretraining: a decoder learns to recover coordinates from the embeddings of a frozen location encoder. (b) A latent diffusion model (\name), conditioned on the image, generates location embeddings, which the frozen LAE decoder maps back to coordinates. We leave out timestep conditioning for visual simplicity.}
    \label{fig:method}
\end{figure}

In this section, we present our generative model, \textbf{D}iffusion for \textbf{I}mage \textbf{G}eolocalization (\textbf{\name}). As seen in Figure~\ref{fig:method}, \name is conditioned on imagery and transports noise to pretrained location embeddings. Using LAEs, generated embeddings are decoded back to original coordinates for prediction.

\subsection{Diffusion for image Geolocalization}

\paragraph{Denoising objective.}
Let $x = E(\ell) \in \mathbb{R}^d$ be the embedding of a location $\ell \in \mathcal{S}^2$
under a location encoder $E$, so that $x \sim p_{\text{data}}$, and let $c$ be
the image condition. We follow rectified flow~\citep{liu2022flow,lipman2022flow} and
define a linear path between noise $\epsilon \sim p_{\text{noise}} = \mathcal{N}(0, I)$ and
data,
\begin{equation}
  z_t = t\,x + (1-t)\,\epsilon, \qquad t \in [0,1], \qquad
  v = \tfrac{\mathrm{d}z_t}{\mathrm{d}t} = x - \epsilon .
\end{equation}
Following JiT~\citep{li2026jit}, our network $\hat{x}_\theta(z_t, t, c)$ directly
predicts the clean embedding ($x$-prediction) rather than the velocity. Unlike JiT which converts this
prediction to a velocity and regresses $v$ ($v$-loss), we follow EDM~\citep{karras2022edm} and keep the loss 
in the prediction space ($x$-loss) and train with the denoising objective:
\begin{equation}
  \mathcal{L}_{\text{denoise}} = \mathbb{E}_{x,\,c,\,t,\,\epsilon}\Big[
  \big\| \hat{x}_\theta(z_t, t, c) - x \big\|^2 \Big] .
\end{equation}

\paragraph{Representation alignment.}
\cite{singh2026raev2} demonstrates how representation alignment (REPA~\citep{yu2024repa}) is
complementary to diffusing in the same representation space for image generation. We apply the same idea and
attach a projection head $h_\phi$ to the features $f_k$ of intermediate block $k$, aligning
them with the location representation $x$:
\begin{equation}
  \mathcal{L}_{\text{REPA}} = -\,\mathbb{E}_{x,\,c,\,t,\,\epsilon}\big[
  \mathrm{sim}\big( h_\phi(f_k),\, x \big) \big] ,
\end{equation}
where $\mathrm{sim}(\cdot,\cdot)$ is a similarity function. The final training loss is
\begin{equation}
  \mathcal{L} = \mathcal{L}_{\text{denoise}} + \lambda\,\mathcal{L}_{\text{REPA}} .
\end{equation}

\paragraph{Sampling and classifier-free guidance.}
Sampling starts from Gaussian noise $z_0 \sim \mathcal{N}(0, I)$, and transports it to a clean embedding by integrating the flow ODE $\mathrm{d}z_t / \mathrm{d}t = v$ from $t = 0$ to $t = 1$. At each step the network predicts the clean embedding $\hat{x}_\theta(z_t, t, c)$, which we convert to a velocity $\hat{v} = \frac{\hat{x}_\theta(z_t, t, c) - z_t}{1 - t},$
and take an Euler step $z_{t+\Delta t} = z_t + \Delta t\,\hat{v}$ over $N$ steps. For
classifier-free guidance \citep{ho2022cfg}, the network is evaluated with and without
the condition and the prediction is extrapolated with guidance scale $s$, 
\begin{equation}
\hat{x} = \hat{x}_\theta(z_t, t, \varnothing) + s\,\big(\hat{x}_\theta(z_t, t, c) - \hat{x}_\theta(z_t, t, \varnothing)\big).
\end{equation}

\subsection{Implementation}

\paragraph{Model architecture.}
Unlike image features, a location representation is a single vector in
$\mathbb{R}^d$. There is no sequence to attend over, so we drop tokenization and attention  altogether and adopt a simpler residual MLP architecture with $D$ blocks and hidden width $w$. 
Following \cite{dufour2025around}, each block is conditioned through adaptive layer normalization (AdaLN). The conditioning vector is the sum of a sinusoidal embedding of $t$ and a linear projection of the image condition $c$. Inspired by MM-DiT~\citep{esser2024sd3}, we build $c$
from two frozen image encoders and concatenate their \texttt{[CLS]} tokens, here StreetCLIP~\citep{haas2023streetclip} and DINOv3~\citep{simeoni2025dinov3}. For
CFG, the condition is replaced by a zero vector with probability
$p_{\text{drop}}=0.1$ during training.

\section{Experiments}
\label{sec:experiments}

\subsection{Experimental Protocol}

\paragraph{Training and inference details.}
We train B and XL models whose hidden widths and depths follow DiT~\citep{peebles2023dit} (Tab.~\ref{tab:sizes}). All models operate on location embeddings from a location encoder $E$, standardized per dimension with training-set statistics, and are conditioned on frozen StreetCLIP and DINOv3 features that are precomputed. The REPA head is attached to block $k = 8$ with $\lambda = 0.5$. We use AdamW~\citep{loshchilov2017decoupled} with learning rate $8\times10^{-4}$ with weight decay $0.05$. Main results use the XL model trained for 1M steps (and an extended 10M step model) following \cite{dufour2025around}. Ablation experiments use the B model trained for 400k steps. All models train on a single NVIDIA H100 with batch size $1024$. At inference we take $50$ Euler steps; guidance scales, step schedule, and hyperparameters are given in App.~\ref{app:ldm:implementation}.

\paragraph{Datasets.}
We evaluate on three image geolocalization benchmarks. \textbf{OSV-5M}~\citep{astruc2024openstreetview} provides 4.9M street-level training images and a 210K-image test set separated from training by a 1~km buffer; we train and test on the official splits. \textbf{Im2GPS3k}~\citep{vo2017revisiting} (2,997 photos) and \textbf{YFCC4k}~\citep{vo2017revisiting} (4K photos from YFCC-100M~\citep{thomee2016yfcc100m}) are Flickr benchmarks. For these we train on MP16-Pro~\citep{jia2024g3}, a 4.1M-image geotagged subset of YFCC-100M. To compare against generative methods trained at scale, we also train on a subset of YFCC-100M, following the filtering protocol from \citet{dufour2025around} and additionally removing test photos of both Flickr benchmarks, yielding 44M images. We denote this subset as YFCC-44M.

\paragraph{Baselines.}
We compare to deterministic~\citep{haas2023streetclip, astruc2024openstreetview, muller2018isn, gadi2026hierloc, weyand2016planet, seo2018cplanet, haas2024pigeon, vivanco2023geoclip, zhou2024img2loc, jia2024g3, jia2025georanker} and generative models~\citep{dufour2025around, wang2025locdiff}. On OSV-5M, baselines are either zero-shot or trained on official splits. For Im2GPS3k and YFCC4k, most baselines train on Flickr-derived geotagged datasets, such as MP-16, MP16-Pro, and YFCC. Full training source breakdown is in Table~\ref{tab:training_sources}. 

\paragraph{Hybrid approach.}
While \name does not use a reference database, it is composable with retrieve-and-rerank
systems~\citep{jia2025georanker}. In this hybrid approach, \name samples serve both as additional candidates and as a prior that rewards nearby database candidates. We apply this to our GeoCLIP model, denoted (+ GeoRanker); see App.~\ref{app:georanker} for details.

\paragraph{Metrics.}
On OSV-5M we follow~\cite{dufour2025around} and report GeoScore~\citep{astruc2024openstreetview}, the mean haversine distance (km), and classification accuracy at country, region and city level. On Im2GPS3k and YFCC4k we report the radius-based accuracy~\cite{vo2017revisiting}, denoting the fraction of predictions within 1, 25, 200, 750 and 2500 kilometers of the ground truth.

\subsection{Main Results}

\setlength{\columnsep}{12pt}
\begin{wraptable}{r}{0.58\textwidth}
\vspace{-14pt}
\centering
\setlength{\tabcolsep}{4pt}
\renewcommand{\arraystretch}{1.1}
\scriptsize
\caption{\textbf{Geolocation performance on OSV-5M}~\citep{astruc2024openstreetview} with official
training and test splits. Dist. is mean distance. \textbf{Bold} indicates best and \textit{italics} second best.}
\label{tab:osv5m_main}
\begin{tabular}{@{}c@{\hspace{5pt}}l ccccc@{}}
\toprule
& & GeoS. $\uparrow$ & Dist. $\downarrow$ & \multicolumn{3}{c}{Accuracy (\%) $\uparrow$} \\
\cmidrule(l){5-7}
& Method & & (km) & Ctry. & Reg. & City \\
\midrule

\multirow{6}{*}{\rotatebox[origin=c]{90}{Deterministic}}
  & SC 0-shot~\citep{haas2023streetclip}           & 2273 & 2854 & 38.4 & 20.8 & 14.8 \\
  & Regression~\citep{astruc2024openstreetview}    & 3028 & 1481 & 56.5 & 16.3 & 0.7  \\
  & ISNs~\citep{muller2018isn}                     & 3331 & 2308 & 66.8 & 39.4 & 4.2  \\
  & Hybrid~\citep{astruc2024openstreetview}        & 3361 & 1814 & 68.0 & 39.4 & 5.9  \\
  & SC Retr.~\citep{haas2023streetclip}            & 3597 & 1386 & 73.4 & 45.8 & 19.9 \\
  & HierLoc~\citep{gadi2026hierloc}                & 3963 & 861  & 82.9 & 55.0 & \textit{23.3} \\
\midrule

\multirow{10}{*}{\rotatebox[origin=c]{90}{Generative}}
  & Uniform~\citep{dufour2025around}               & 131  & 10052 & 2.4  & 0.1  & 0.0  \\
  & vMFMix~\citep{izbicki2019exploiting}           & 1746 & 5662  & 34.2 & 11.1 & 0.3  \\
  & vMF~\citep{dufour2025around}                   & 2776 & 2439  & 52.7 & 17.2 & 0.6  \\
  & FM $\mathbb{R}^3$~\citep{dufour2025around}     & 3688 & 1149  & 74.9 & 40.0 & 4.2  \\
  & Diff $\mathbb{R}^3$~\citep{dufour2025around}   & 3762 & 1123  & 75.9 & 40.9 & 3.6  \\
  & RFM $\mathcal{S}_2$~\citep{dufour2025around}   & 3767 & 1069  & 76.2 & 44.2 & 5.4  \\
  & LocDiff~\citep{wang2025locdiff}                & --   & --    & 77.0 & 46.3 & 11.0 \\
 &  \multicolumn{6}{@{}l}{\hl \textbf{\textit{\name-XL (ours)}}} \\
  & GeoCLIP & 4095 & \textit{851} & \textit{83.4} & \textit{60.1} & \textbf{24.6} \\
  & SatCLIP & \textit{4125} & 912 & \textit{83.4} & \textbf{61.0} & 22.4 \\
  & TTE     & \textbf{4129} & \textbf{849} & \textbf{83.5} & 57.4 & 18.4 \\
\bottomrule
\end{tabular}
\vspace{-14pt}
\end{wraptable}

\paragraph{Comparison to generative methods.}
In Table~\ref{tab:osv5m_main}, all \name-XL models surpass prior generative methods on every metric. The gains grow at finer scales, with region accuracy improving by 11--15 points and city accuracy doubling.

In Table~\ref{tab:im2gps_yfcc_main}, we compare on Im2GPS3k and YFCC4k. On YFCC4k, each generative baseline trains on the 48M-image YFCC subset, whereas our models train on MP16-Pro (4.1M) or on YFCC-44M. Trained on ${\sim}10\times$ less data, our MP16-Pro GeoCLIP model already matches or exceeds RFM$_{10\text{M}}$ at 200--2500\,km, and surpasses it by 7.7 points at 1\,km. Trained on YFCC-44M at a matched 1M steps, our GeoCLIP model outperforms all generative methods at each metric. Notably, scaling to 10M steps widens the margin over RFM$_{10\text{M}}$ to 32.1 points at 1\,km.

On Im2GPS3k, Diff~$\mathbb{R}^3$, FM~$\mathbb{R}^3$ and RFM$_{10\text{M}}$ train on YFCC-48M without filtering against the test set, which may inflate their scores. Our YFCC-44M GeoCLIP model nevertheless exceeds them at every threshold, and scaling to 10M steps further widens the gap. While LocDiff and LocDiff-H train on MP16 and use 16 augmented forward passes, our models use a single forward pass trained on the smaller MP16-Pro. Despite this, our GeoCLIP model is ahead at 1 and 25\,km and within 3.2 points at 200--2500\,km. Finally, the TTE and SatCLIP models match or exceed GeoCLIP at 200--2500\,km but fall well behind at 1 and 25\,km. We return to this trade-off in Sec.~\ref{sec:discussion}.

\begin{table*}[t]
\centering
\setlength{\tabcolsep}{2.5pt}
\renewcommand{\arraystretch}{1.0}
\scriptsize
\caption{\textbf{Geolocation performance on Im2GPS3k and YFCC4k.} Accuracy (\%) within
distance thresholds, with best results in \textbf{bold}. We train different \name-XL variants on MP16-Pro and YFCC-44M. 10M denotes training for 10M steps; $^\ddag$~denotes reimplementation with public weights.}
\label{tab:im2gps_yfcc_main}
\begin{tabular}{@{}c@{\hspace{4pt}}l ccccc | ccccc@{}}
\toprule
& & \multicolumn{5}{c}{Im2GPS3k} & \multicolumn{5}{|c}{YFCC4k} \\
\cmidrule(lr){3-7}\cmidrule(l){8-12}
& Method
  & 1\,km & 25\,km & 200\,km & 750\,km & 2500\,km
  & 1\,km & 25\,km & 200\,km & 750\,km & 2500\,km \\
\midrule

\multirow{10}{*}{\rotatebox[origin=c]{90}{Deterministic}}
  & PlaNet~\citep{weyand2016planet}                   & 8.5  & 24.8 & 34.3 & 48.4 & 64.6 & 5.6  & 14.3 & 22.2 & 36.4 & 55.8 \\
  & CPlaNet~\citep{seo2018cplanet}                  & 10.2 & 26.5 & 34.6 & 48.6 & 64.6 & 7.9  & 14.8 & 21.9 & 36.4 & 55.5 \\
  & Translocator~\citep{pramanick2022translocator}             & 11.8 & 31.1 & 46.7 & 58.9 & 80.1 & 8.4  & 18.6 & 27.0 & 41.1 & 60.4 \\
  & GeoDecoder~\citep{clark2023geodecoder}               & 12.8 & 33.5 & 45.9 & 61.0 & 76.1 & 10.3 & 24.4 & 33.9 & 50.0 & 68.7 \\
  & PIGEOTTO~\cite{haas2024pigeon}                 & 11.3 & 36.7 & 53.8 & 72.4 & 85.3 & 10.4 & 23.7 & 40.6 & 62.2 & 77.7 \\
  & GeoCLIP~\citep{vivanco2023geoclip}                  & 14.1 & 34.5 & 50.7 & 69.7 & 83.8 & 9.6  & 19.3 & 32.6 & 55.0 & 74.7 \\
  & HierLoc~\citep{gadi2026hierloc}                  & 11.3 & 43.8 & 58.4 & 74.1 & 85.1 & 8.4  & 30.2 & 43.3 & 61.7 & 75.8 \\
  & Img2Loc~\citep{zhou2024img2loc}                  & 17.1 & 45.1 & 57.9 & 72.9 & 84.7 & 14.1 & 29.6 & 41.4 & 59.3 & 76.9 \\
  & G3~\citep{jia2024g3}                       & 16.7 & 40.9 & 55.6 & 71.2 & 84.7 & 24.0 & 35.9 & 47.0 & 64.3 & 78.2 \\
  & GeoRanker~\citep{jia2025georanker}                & 18.8 & 45.1 & 61.5 & 76.3 & 89.3 & 32.9 & 43.5 & 54.3 & 69.8 & 82.5 \\
\midrule

\multirow{18}{*}{\rotatebox[origin=c]{90}{Generative}}
  & vMFMix~\citep{izbicki2019exploiting}                                    & --   & --   & --   & --   & --   & --   & 0.4  & 8.8  & 20.9 & 41.0 \\
  & vMF~\citep{dufour2025around}                                       & --   & --   & --   & --   & --   & --   & 4.8  & 15.0 & 30.9 & 53.4 \\
  & Diff $\mathbb{R}^3$~\citep{dufour2025around}                       &  0.0\rlap{$^\ddag$} & 16.2\rlap{$^\ddag$} & 46.8\rlap{$^\ddag$} & 63.1\rlap{$^\ddag$} & 79.1\rlap{$^\ddag$} &  0.0\rlap{$^\ddag$} & 11.1 & 37.7 & 54.7 & 71.9 \\
  & FM $\mathbb{R}^3$~\citep{dufour2025around}                         &  2.1\rlap{$^\ddag$} & 29.1\rlap{$^\ddag$} & 43.8\rlap{$^\ddag$} & 60.8\rlap{$^\ddag$} & 78.6\rlap{$^\ddag$} &  2.0\rlap{$^\ddag$} & 22.1 & 35.0 & 53.2 & 73.1 \\
  & RFM $\mathcal{S}_2$~\citep{dufour2025around}                       & --   & --   & --   & --   & --   & --   & 23.7 & 36.4 & 54.5 & 73.6 \\
  & RFM\textsubscript{10M} $\mathcal{S}_2$~\citep{dufour2025around}    &  8.2\rlap{$^\ddag$} & 37.2\rlap{$^\ddag$} & 51.3\rlap{$^\ddag$} & 66.8\rlap{$^\ddag$} & 81.7\rlap{$^\ddag$} &  9.8\rlap{$^\ddag$} & 33.5 & 45.3 & 61.1 & 77.7 \\
  & LocDiff~\citep{wang2025locdiff}                          & 10.9 & 34.0 & 53.3 & 72.5 & 85.2 & --   & 33.3 & 46.7 & 65.2 & 79.9 \\
  & LocDiff-H~\citep{wang2025locdiff}                                 & 15.3 & 36.5 & 56.4 & 75.2 & 87.4 & --   & --   & --   & --   & --   \\
    & \multicolumn{11}{@{}l}{\hl \textit{\textbf{\name-XL (ours)}, trained on MP16-Pro}} \\
  & TTE                        & 3.9 & 34.5 & 56.3 & 73.0 & 85.5 & 1.8 & 23.4 & 47.7 & 65.0 & 77.6 \\
  & SatCLIP                    & 2.4 & 36.4 & 55.5 & 72.0 & 84.8 & 1.7 & 26.2 & 47.7 & 64.3 & 78.9 \\
  & GeoCLIP                    & 17.3 & 41.6 & 56.2 & 72.0 & 86.3 & 17.5 & 31.6 & 45.0 & 63.4 & 78.8 \\
  & GeoCLIP + GeoRanker        & 20.1 & 47.6 & \textbf{64.1} & \textbf{78.1} & \textbf{89.6} & 34.0 & 45.5 & 56.8 & 71.9 & 83.8 \\
    & \multicolumn{11}{@{}l}{\hl \textit{\textbf{\name-XL (ours)}, trained on YFCC-44M}} \\
  & TTE                         & 2.2 & 36.0 & 59.6 & 74.9 & 87.3 & 0.9 & 30.1 & 55.3 & 70.8 & 82.7 \\
  & SatCLIP                     & 3.4 & 39.3 & 60.3 & 75.4 & 87.4 & 2.7 & 36.6 & 58.4 & 72.0 & 82.7 \\
  & GeoCLIP                     & 20.2 & 45.4 & 58.4 & 73.6 & 86.8 & 30.1 & 47.5 & 58.3 & 71.1 & 82.9 \\
  & GeoCLIP\textsubscript{10M} & \textbf{21.9} & \textbf{48.6} & 60.8 & 75.3 & 87.6 & \textbf{41.9} & \textbf{59.7} & \textbf{67.9} & \textbf{77.1} & \textbf{86.1} \\
\bottomrule
\end{tabular}
\end{table*}

\paragraph{Comparison to deterministic methods.}
In Table~\ref{tab:osv5m_main}, our GeoCLIP model surpasses the strongest deterministic method, HierLoc, on every metric. The SatCLIP and TTE models trail HierLoc at city level while exceeding it on GeoScore, country and region accuracy, and TTE also on mean error. Retrieval-based methods are inherently limited on OSV-5M, whose test images are separated from the training set by a spatial buffer, so no gallery entry is at the true location. Our generative model predicting in continuous space does not have this ceiling.

On Im2GPS3k and YFCC4k (Table~\ref{tab:im2gps_yfcc_main}), most deterministic methods also train on MP16 (Table~\ref{tab:training_sources}). On Im2GPS3k, our MP16-Pro GeoCLIP model is second best at 1 and 2500\,km and within 3.5--5.3 points of the best method elsewhere. On YFCC4k the gap to deterministic methods is larger at fine scales and closes at coarse ones. We attribute this to the geographic overlap between MP16 and the benchmarks, which retrieval exploits directly (App.~\ref{app:dataset_comparison}). We leverage this by composing \name with GeoRanker, a strong retrieve-and-rerank method, which outperforms every deterministic method on both benchmarks, including GeoRanker itself. Scaling the training data has a similar effect. Our YFCC-44M GeoCLIP model trained for 10M steps surpasses all deterministic methods on YFCC4k and at 1 and 25\,km on Im2GPS3k.

\begin{figure}[H]
    \centering
    \includegraphics[width=\textwidth]{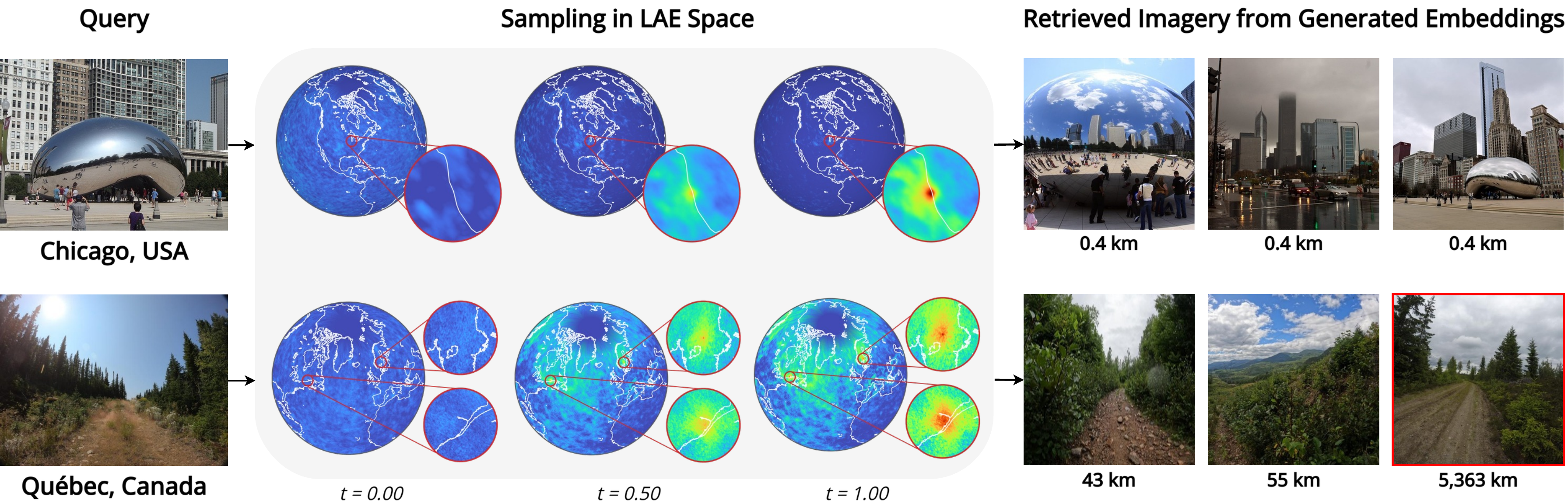}
    \caption{\textbf{Sampling in the LAE space.} For each query \textit{(left)}, we show three sampling trajectories as the cosine similarity between the latents and location embeddings across the globe \textit{(middle)}, and display the top images retrieved with the generated embeddings, including their error in km \textit{(right)}.
}
    \label{fig:main_qual}
\end{figure}

\paragraph{Sampling in a semantic space.}
Figure~\ref{fig:main_qual} shows two test queries. The Cloud Gate sculpture in Chicago
(top) is highly localizable~\citep{astruc2024openstreetview, izbicki2019exploiting}, and
the cosine similarity between the sampled latent and location embeddings across the globe
resolves to a sharp peak. The generated embeddings retrieve images of the
sculpture itself or of the Chicago skyline from within 0.4~km, showing that they carry
semantics and position aligned with the query. The forest track near Baie-Saint-Paul, Qu\'ebec
(bottom) instead offers few distinctive cues, and its boreal conifer forest also
characterizes southern Scandinavia. The similarity map is correspondingly diffuse,
concentrating on both regions, and one trajectory commits to the wrong continent. Yet
its retrieval is a similar forest road, demonstrating that the error is semantically intuitive.
We attribute this behavior to diffusing in a representation space where distant but
semantically similar places lie close together. In App.~\ref{app:posterior}, we exploit
this structure by drawing multiple samples and predicting from the densest mode, which
improves accuracy.

\subsection{\name Design Choices}

\newcommand{\ablhead}[1]{%
  \toprule
  #1 & GeoS. $\uparrow$ & Dist. & \multicolumn{3}{c}{Acc. (\%) $\uparrow$} \\
  \cmidrule(l){4-6}
  & & (km) $\downarrow$ & Ctry. & Reg. & City \\
  \midrule
}

\newcommand{\panelgap}{\par\vspace{4pt}}   %
\newcommand{\capgap}{\vspace{2pt}}          %

\begin{wraptable}{r}{0.4\textwidth}
\vspace{-14pt}
\centering
\setlength{\tabcolsep}{2.5pt}
\renewcommand{\arraystretch}{1.1}
\scriptsize

\caption{\textbf{\name ablations.} We train \name-B models on OSV-5M using GeoCLIP LAEs; the {\setlength{\fboxsep}{1pt}\colorbox{cyan!10}{highlighted}} row in each panel is our chosen configuration.}
\label{tab:ablations}

\setcounter{subtable}{0}
\begin{subtable}[t]{\linewidth}
\centering
\scriptsize
\begin{tabular}{@{}lccccc@{}}
\ablhead{Method}
$\epsilon$-pred, $\epsilon$-loss & 3847 & 1007 & 79.3 & 47.2 & 11.6 \\
$v$-pred, $v$-loss               & 3968 & 920 & 81.5 & 53.1 & 16.4 \\
$x$-pred, $\epsilon$-loss        & 3795 & 1053 & 78.4 & 45.4 & 10.5 \\
$x$-pred, $v$-loss               & 3957 &  947 & 80.8 & 53.6 & 17.4 \\
\rowcolor{cyan!10} $x$-pred, $x$-loss & \textbf{4037} & \textbf{878} & \textbf{82.6} & \textbf{56.4} & \textbf{19.8} \\
\bottomrule
\end{tabular}
\capgap
\caption{\textbf{$\bm{x}$ marks the prediction and loss.}}
\label{tab:abl_param}
\end{subtable}

\panelgap

\setcounter{subtable}{1}
\begin{subtable}[t]{\linewidth}
\centering
\scriptsize
\begin{tabular}{@{}lccccc@{}}
\ablhead{Schedule}
linear       & 3967 & 951 & 81.1 & 54.3 & 18.7 \\
skew-sigmoid & 3986 & 923 & 82.2 & 54.3 & 17.9 \\
\rowcolor{cyan!10} sigmoid & \textbf{4037} & \textbf{878} & \textbf{82.6} & \textbf{56.4} & \textbf{19.8} \\
\bottomrule
\end{tabular}
\capgap
\caption{\textbf{Diffusion schedule choice.}}
\label{tab:abl_schedule}
\end{subtable}

\panelgap

\setcounter{subtable}{2}
\begin{subtable}[t]{\linewidth}
\centering
\scriptsize
\begin{tabular}{@{}lccccc@{}}
\ablhead{Method}
w/o REPA & 3999 & 903 & 81.8 & 54.0 & 16.9 \\
\rowcolor{cyan!10} w/ REPA & \textbf{4037} & \textbf{878} & \textbf{82.6} & \textbf{56.4} & \textbf{19.8} \\
\bottomrule
\end{tabular}
\capgap
\caption{\textbf{REPA improves performance.}}
\label{tab:abl_repa}
\end{subtable}

\panelgap

\setcounter{subtable}{3}
\begin{subtable}[t]{\linewidth}
\centering
\scriptsize
\begin{tabular}{@{}lccccc@{}}
\ablhead{Encoder}
DINOv3     & 3824 & 1206 & 77.7 & 52.1 & 17.5 \\
StreetCLIP & 3953 & 975 & 81.0 & 53.8 & 18.2 \\
\rowcolor{cyan!10} Both & \textbf{4037} & \textbf{878} & \textbf{82.6} & \textbf{56.4} & \textbf{19.8} \\
\bottomrule
\end{tabular}
\capgap
\caption{\textbf{Combining vision representations.}}
\label{tab:abl_encoder}
\end{subtable}

\vspace{-20pt}
\end{wraptable}

\paragraph{Diffusion ablations.}
Table~\ref{tab:ablations} ablates the diffusion recipe with \name-B on OSV-5M. \emph{(a)~Prediction and loss space.} Score matching ($\epsilon$-pred, $\epsilon$-loss) performs poorly, while flow matching ($v$-pred, $v$-loss) improves but still lags at fine scales (16.4\% at city level). Predicting and regressing the clean embedding ($x$-pred, $x$-loss) is best on every metric. \emph{(b)~Schedule.} \citet{dufour2025around} use a skewed sigmoid that concentrates training on low-noise timesteps near the data. While this skewed sigmoid beats a linear schedule on GeoScore and mean error in our LAE diffusion space, it trails at city level (17.9 vs.\ 18.7). A symmetric sigmoid, which emphasizes both ends of the path equally, is best on all metrics. \emph{(c)~REPA.} Similar to \citet{singh2026raev2}, we find that representation alignment is complementary to diffusing in the same representation space. REPA adds 38 GeoScore and 2.9 points of city accuracy. \emph{(d)~Conditioning.} Concatenating StreetCLIP and DINOv3 features outperforms either alone. We hypothesize this is due to StreetCLIP carrying geolocation-specific discriminative features while DINOv3 supplying general visual features, both beneficial for the diffusion model.

\begin{figure}[H]
\centering
\begin{minipage}[t]{0.5\textwidth}
    \centering
    \setlength{\tabcolsep}{3pt}
    \renewcommand{\arraystretch}{1.1}
    \scriptsize
    \captionof{table}{\textbf{Importance of semantic latent spaces.} We train \name-B models evaluated at OSV-5M.}
    \label{tab:lae_importance}
    \vspace{-8pt}
    \begin{tabular}{@{}lcccccc@{}}
    \toprule
    Diffusion Space & GeoS. $\uparrow$ & \multicolumn{2}{c}{Dist. (km) $\downarrow$} & \multicolumn{3}{c}{Classification Acc. (\%) $\uparrow$} \\
    \cmidrule(lr){3-4} \cmidrule(l){5-7}
    & & Mean & Median & Country & Region & City \\
    \midrule
    coordinates  & 3425 & 1604 & 199.0 & 68.7 & 38.8 & 6.0 \\
    SH ($L=10$)  & 4042 & \textbf{825} & 116.9 & 81.0 & 47.9 & 9.1 \\
    SH ($L=20$)  & 4051 & 856 & 101.9 & 81.5 & 50.6 & 10.7 \\
    SH ($L=40$)  & 4023 & 979 & 83.4 & 81.1 & 53.8 & 13.9 \\
    \midrule
    \rowcolor{cyan!10} \multicolumn{7}{@{}l}{\textit{LAE (ours)}} \\
    SatCLIP      & \textbf{4096} & 875 & \textbf{65.6} & \textbf{82.9} & \textbf{57.6} & \textit{18.4} \\
    TTE          & \textit{4089} & \textit{847} & 82.0 & 82.5 & 54.1 & 15.4 \\
    GeoCLIP      & 4037 & 878 & \textit{70.9} & \textit{82.6} & \textit{56.4} & \textbf{19.8} \\
    \bottomrule
    \end{tabular}
\end{minipage}%
\hfill
\begin{minipage}[t]{0.44\textwidth}
    \centering
    \vspace{0pt}
    \includegraphics[width=0.92\linewidth]{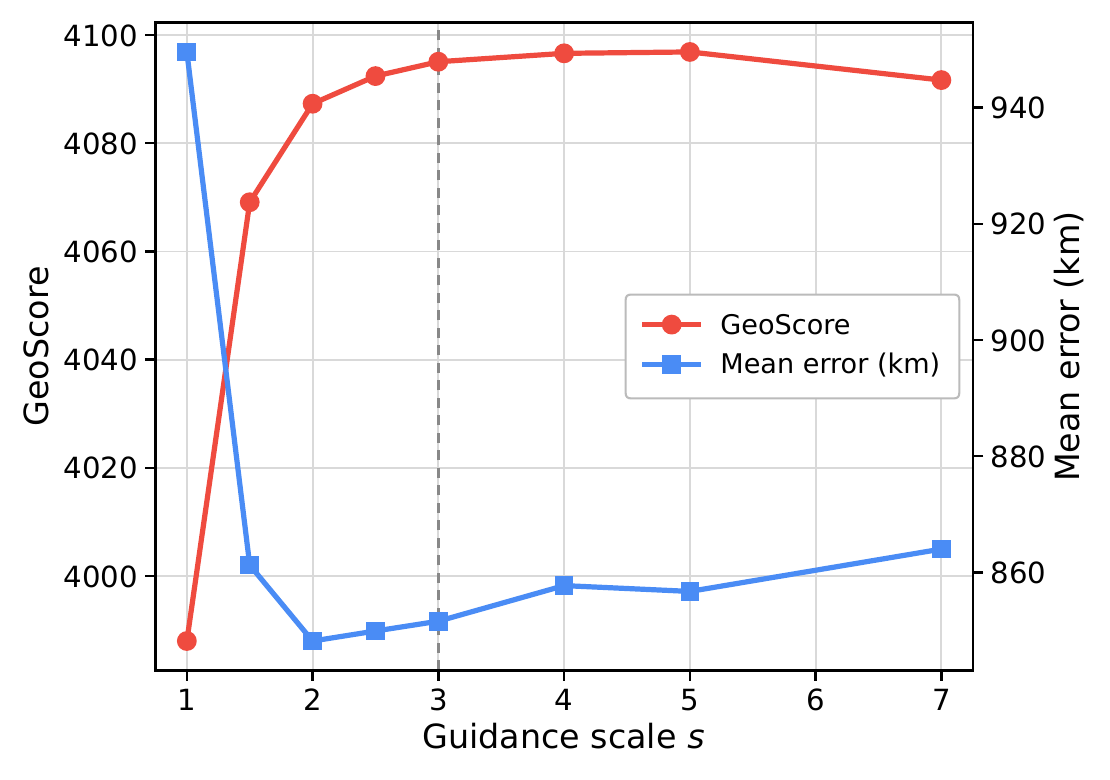}
    \vspace{-9pt}
    \caption{\textbf{Impact of CFG.} \name-XL with GeoCLIP LAE evaluated on OSV-5M.}
    \label{fig:cfg_sweep}
\end{minipage}
\end{figure}

\paragraph{Importance of pretrained location representations.}
Table~\ref{tab:lae_importance} varies the diffusion space, using \name-B models on OSV-5M. Diffusing directly in  coordinate space performs worst. SH spaces perform far better, with city-level performance increasing with spectral resolution ($L$), demonstrating the value of multi-scale positional information~\citep{wang2025locdiff}. Yet, SatCLIP, which learns a 256-dimensional projection from the SH ($L=40$) encoding, improves over it on every metric, as do TTE and GeoCLIP likewise. Despite their lower dimensionality, learned representations that add semantic structure make for a better diffusion space.

\paragraph{Classifier-free guidance.}
Consistent with \citet{dufour2025around}, guidance improves geolocalization performance (Figure~\ref{fig:cfg_sweep}). Without CFG ($s=1$), GeoScore drops and mean error rises. Gains saturate by $s \approx 3$, so we use $s=3$ throughout.

\section{Discussion}
\label{sec:discussion}

\paragraph{Location encoder trade-offs.}
GeoCLIP is strongest at fine scales and SatCLIP and TTE at coarse ones, but the latter trail at 1--25\,km, especially on Flickr benchmarks. We see two likely reasons:

\emph{(1) Alignment modality.} SatCLIP and TTE are aligned to Sentinel-2 satellite imagery~\citep{klemmer2025satclip}, whereas GeoCLIP is aligned to ground-level photographs~\citep{larson2017benchmarking}. \name depends on the \emph{semantics} these encoders carry (Table~\ref{tab:lae_importance}), so those semantics should ideally match the query imagery. On OSV-5M street views, the encoder gap is smaller as overhead appearance corresponds reasonably well to street-level semantics. On the Flickr benchmarks it is large, since a satellite view of Paris says little about a caf\'e menu or a museum hall photographed there, making the these encoders contribute little beyond position. 

\emph{(2) Spectral resolution.} GeoCLIP's random Fourier features encode much higher spatial frequencies than the band-limited spherical harmonics of SatCLIP or TTE's soft site assignment, making its space more locally discriminative, accounting for its fine-grained accuracy. We see the same trend in Table~\ref{tab:lae_importance}, where a higher Legendre degree $L$, and thus finer spectral resolution, improves city-level accuracy. Conversely, the global bases and semantic tokens of SatCLIP and TTE capture structure at continental scale, which explains their higher accuracy at coarse scales. Leveraging an ensemble of location encoders or ones aligned to multiple modalities, is a natural future direction.

\paragraph{Sub-kilometer accuracy.}
\begin{wraptable}{r}{0.38\linewidth}
    \centering
    \vspace{-\intextsep}
    \scriptsize
    \setlength{\tabcolsep}{2.5pt}
    \caption{\textbf{Sub-km reconstruction.}}
    \label{tab:sub_km}
    \vspace{-8pt}
    \begin{tabular}{@{}lccc|c@{}}
        \toprule
        Decoder & $<$0.01\,km & $<$0.1\,km & $<$0.5\,km & GFLOPs \\
        \midrule
        SatCLIP & 0.10 & 9.31 & 89.0 & 0.076 \\
        GeoCLIP & 0.01 & 1.23 & 26.1 & 0.076 \\
        TTE     & 0.09 & 8.70 & 85.6 & 0.076 \\
        \midrule
        \rowcolor{cyan!10} \multicolumn{5}{@{}l}{\textit{+ 5 refinement steps}} \\
        SatCLIP & 90.2 & 99.4 & 100.0 & 0.161 \\
        GeoCLIP & 99.7 & 99.8 & 99.9 & 0.737 \\
        TTE     & 100.0 & 100.0 & 100.0 & 0.315 \\
        \bottomrule
    \end{tabular}
\end{wraptable}

Our decoders target reconstruction accuracy at the kilometer-level, matching the scale of global geolocalization. Yet many tasks, such as trajectory geolocalization~\citep{gao2026trajloc}, operate below 1\,km. We probe reconstruction at this scale in Table~\ref{tab:sub_km}, finding raw decoding to be poor. Analytic refinement resolves this for GeoCLIP and TTE, while SatCLIP plateaus regardless of refinement. Matching location encoders to higher-resolution geolocalization tasks remains open.

\section{Conclusion}
\label{sec:conclusion}
In this work, we proposed \name, a diffusion model for image geolocalization that leverages pretrained location encoders by diffusing in their representation space. Geolocating a photograph requires understanding both where a place is and its semantic context, which location encoders capture together in a single embedding. Instead of predicting the image's coordinates directly, we propose to generate their location embeddings with \name, a conditional latent diffusion model. To use these generating representations for prediction, we introduce a Location Autoencoder (LAE) that decodes them back to original coordinates. Using this recipe, \name sets a new state of the art on OSV-5M, Im2GPS3k and YFCC4k, and we demonstrate that its performance emerges from using location encoders that learn semantics beyond their positional encoding. Notably, \name can be composed with retrieve-and-rerank systems to improve their performance at every distance threshold. Location encoders were built to describe the Earth; we show they are also the space in which to search it.

\section*{Acknowledgments}
This research used the TGI RAILs advanced compute and data resource, which is supported by the National Science Foundation (award OAC-2232860) and the Taylor Geospatial Institute. This work is also supported by the AI-ACCESS National Research Traineeship, funded by the National Science Foundation (award DGE-2244165).

\subsection*{AI use statement}

In this work, we used generative AI tools to help develop theoretical models
and conceptual frameworks, to design and provide feedback on our research
methodology and experiments, to implement methods, and to clean and reformat
datasets. We have not used generative AI tools to propose or refine
hypotheses, to interpret results, to support qualitative or thematic data
analysis, or to assist with translation. Generating synthetic datasets,
formulating mathematical claims, providing critical ingredients for proving
mathematical claims, and assisting in the writing of proofs are not applicable
to this work. Additionally, we used generative AI tools to refine the writing
of sections of this paper and to help identify related literature. We have
reviewed all AI-assisted work. All AI-generated code was reviewed and
verified for correctness by the authors, and all AI-suggested references were
checked against the original sources. We take responsibility
for the final content of this work, including text, claims, or artifacts
produced with the aid of generative AI.

\subsection*{Ethics statement}

In this work, we study how pretrained
representations can be leveraged within generative models for
geolocalization. All experiments use established public datasets and benchmarks,
namely MP16-Pro, OSV-5M, YFCC4K, Im2GPS3k, and YFCC-100M, each of which
we cite from its original source and use in accordance with its
research license. We did not collect any new data, and no human
subjects were involved in this research. We note that these datasets contain user-uploaded photographs depicting people or private settings. We use these images as released and do not attempt to identify individuals.
More broadly, we recognize that improvements in geolocalization carry
risk. Geolocalization can enable nonconsensual tracking and expose private residences, personal routines, or sensitive locations. Our models are intended for research and benchmarking, and we do not endorse their use to surveil or track individuals. Performance varies geographically, with potentially larger errors in regions underrepresented in training data. These disparities and privacy risks warrant careful consideration before deployment.

\bibliography{iclr2027_conference}
\bibliographystyle{iclr2027_conference}

\newpage
\appendix
\section{Appendix}

\subsection{Location Autoencoders}
\label{app:lae}

\subsubsection{Location Encoders}
We evaluate spherical harmonic (SH) encodings~\citep{russwurm2024geographic}
at three truncation levels, $L \in \{10,20,40\}$, and three pretrained location encoders: SatCLIP~\citep{klemmer2025satclip},
TTE~\citep{cher2026tte}, and GeoCLIP~\citep{vivanco2023geoclip}.
The pretrained encoders use contrastive learning to align geographic coordinates with features of images captured at those locations.

\paragraph{Spherical harmonics.}
Spherical harmonics provide a Fourier-like basis on the sphere.
Following \citet{russwurm2024geographic}, we encode latitude $\varphi$
and longitude $\lambda$ by evaluating the real spherical harmonics
of degrees $0,\ldots,L-1$:
\begin{equation}
\mathrm{PE}_L(\varphi,\lambda)
=
\big[\,Y_l^m(\varphi,\lambda)\,\big]_{0 \le l < L,\; |m| \le l}
\in \mathbb{R}^{L^2},
\end{equation}
where
\begin{equation}
Y_l^m(\varphi,\lambda)
=
N_l^m P_l^{|m|}(\sin\varphi)
\begin{cases}
\cos(m\lambda), & m > 0,\\
1, & m = 0,\\
\sin(|m|\lambda), & m < 0.
\end{cases}
\end{equation}
Here, $P_l^{|m|}$ is an associated Legendre polynomial, and $N_l^m$
normalizes each basis function to unit norm on the sphere.
Larger $L$ includes higher-frequency harmonics and supports finer
spatial detail.
The encoding has $L^2$ dimensions: 100, 400, and 1,600 for
$L=10$, $20$, and $40$, respectively.
We use these fixed features directly as the diffusion latent.

\paragraph{Comparing representations.}
\begin{figure}[H]
    \centering
    \includegraphics[width=\textwidth]{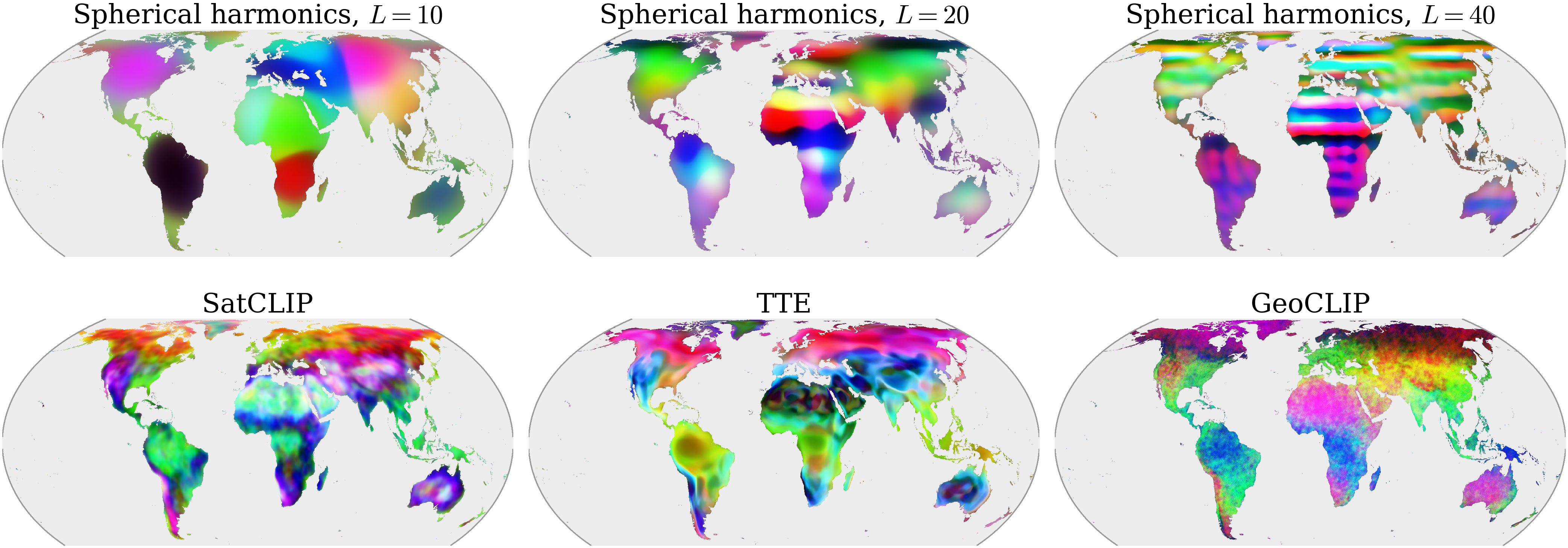}
    \caption{\textbf{ICA of location embeddings over land.} Top: spherical harmonics with $L{=}10, 20, 40$. Bottom: learned encoders SatCLIP, TTE, and GeoCLIP. Each panel maps its first three independent components to RGB.}
    \label{fig:ica_maps}
\end{figure}

We visualize each encoder with independent component analysis (ICA) in Figure~\ref{fig:ica_maps}, following \citet{dhakal2025range}: embeddings are evaluated over land, standardized, and reduced to three independent components shown as RGB. Colors are arbitrary per panel, so only spatial structure is comparable. For SH, increasing $L$ raises the spatial frequency, from continent-scale blobs at $L{=}10$ to regular banding along parallels at $L{=}40$. SatCLIP's SirenNet maps the same $L{=}40$ basis to smooth regions that follow geographic structure rather than parallels, and has been shown to better capture location semantics~\citep{klemmer2025satclip}. TTE's map is smooth, since the soft assignment interpolates between neighboring sites. GeoCLIP's multi-scale RFF positional encoding produces a much higher-frequency map than the other encoders.

\subsubsection{LAE Implementation Details}
\label{app:lae_implementation}

\paragraph{Decoder architecture.}
All decoders share one design. The encoder's raw output ($d{=}512$ for GeoCLIP and TTE, $d{=}256$ for SatCLIP, and $d{=}L^2$ for SH) is layer-normalized and projected to width $w$. It is then processed by 8 pre-norm residual MLP blocks. Each block applies LayerNorm, a linear layer from $w$ to $4w$, GELU, and a linear layer from $4w$ back to $w$. The last layer of each block is zero-initialized so that every block starts as the identity. A final LayerNorm and linear layer map the features to $\mathbb{R}^3$, and the output is normalized onto the unit sphere. We use $w{=}768$.

\paragraph{Decoder training.}
Decoders are trained only on synthetic pairs. Each step draws a fresh batch of 4096 locations uniformly by area over land, using rejection sampling against a $0.05^\circ$ raster of the 1:10m Natural Earth land and minor-island polygons\footnote{\url{https://www.naturalearthdata.com}}. The frozen encoder embeds these locations, and the decoder minimizes the mean great-circle distance in kilometers between decoded and true locations. We use Adam~\citep{kingma2014adam} ($\beta_1{=}0.9$, $\beta_2{=}0.999$) with a constant learning rate of $3{\times}10^{-4}$ after a 1k-step linear warm-up, and clip gradients at norm 1.0. Training runs on a single H100 GPU. We evaluate an exponential moving average of the weights with decay 0.999. Locations are resampled at every step, so no location is repeated and no decoder is fit to any benchmark's coordinate distribution. We measure reconstruction on a fixed set of 1M land locations drawn the same way with a separate seed. Table~\ref{tab:decoder_training} lists the training length for each encoder.
\begin{table}[H]
\centering
\caption{\textbf{Official decoder training.} All decoders have depth 8 and use batch size 4096, with locations drawn uniformly over land. Density is locations seen per km$^2$ of land ($1.50\times10^{8}$\,km$^2$).}
\label{tab:decoder_training}
\small
\begin{tabular}{@{}lccccc@{}}
\toprule
Encoder & Input dim & Params & Steps & Locations seen & Density (/km$^2$) \\
\midrule
GeoCLIP       & 512   & 38.2M  & 6.1M & $2.50\times10^{10}$ & 167 \\
TTE           & 512   & 38.2M  & 1.5M & $6.14\times10^{9}$  & 41  \\
SatCLIP       & 256   & 38.0M  & 400k & $1.64\times10^{9}$  & 11  \\
\midrule
SH ($L{=}10$) & 100  & 37.9M & 200k & $8.19\times10^{8}$ & 5.5 \\
SH ($L{=}20$) & 400  & 38.1M & 200k & $8.19\times10^{8}$ & 5.5 \\
SH ($L{=}40$) & 1600 & 39.0M & 200k & $8.19\times10^{8}$ & 5.5 \\
\bottomrule
\end{tabular}
\end{table}

\paragraph{Levenberg--Marquardt refinement.}
We refine each decoded location $\hat{\ell}_0 = D_\psi(x)$ by minimizing the encoder residual $r(\ell) = x - E(\ell)$ with Equation~\ref{eq:lm}. Updates are taken in a local tangent frame at $\hat{\ell}_k$, with unit east and north vectors $u_e$ and $u_n$ and steps measured in kilometers. The Jacobian $J_k \in \mathbb{R}^{d\times 2}$ is computed exactly with two forward-mode automatic differentiation passes, one along $u_e$ and one along $u_n$. Solving the damped normal equations gives a step $\delta_k = (\delta_e, \delta_n) \in \mathbb{R}^2$, and $\oplus$ follows the great circle in that direction,
\begin{equation}
\hat{\ell} \oplus \delta = \cos\!\Big(\frac{\|\delta\|}{R}\Big)\,\hat{\ell} + \sin\!\Big(\frac{\|\delta\|}{R}\Big)\,\frac{\delta_e u_e + \delta_n u_n}{\|\delta\|},
\end{equation}
where $R = 6371$\,km is the Earth's radius. A step is accepted only if it lowers $\|r\|$, and a rejected step leaves the location unchanged. Each location keeps its own damping $\lambda_k$, starting at $\lambda_0 = 10^{-3}$. It is divided by 3 after an accepted step and multiplied by 10 after a rejected one. Each step costs two forward-mode passes, one encoder evaluation for the acceptance check, and a batched $2{\times}2$ solve, and the procedure is deterministic. We use five steps for GeoCLIP, one for SatCLIP and TTE.

\subsubsection{Additional LAE Experiments}

\paragraph{Analytic refinement.}
We vary the number of LM steps and measure the effect on reconstruction (Table~\ref{tab:lae_refine_all}) and on geolocalization (Table~\ref{tab:refine_benchmarks}). For reconstruction, a single step raises the GeoCLIP decoder from 67.91\% to 99.10\% within 1\,km, and five steps reach 99.89\%. The SatCLIP and TTE decoders start at about 99.6\% and reach 100.00\% after one step. Cost grows linearly with the number of steps. Decoding with five GeoCLIP steps costs 0.74 GFLOPs per location, under 1\% of the roughly 84 GFLOPs needed to sample one latent with \name-XL (50 steps with CFG). We therefore use five steps for GeoCLIP and one step for SatCLIP and TTE.

Table~\ref{tab:refine_benchmarks} decodes the same generated samples with 0 to 5 steps. On every benchmark, refinement changes accuracy at 25\,km and above by at most 0.3 points, so its effect is limited to the 1\,km threshold. On OSV-5M, fewer than 0.3\% of predictions fall within 1\,km for any encoder or step count. On the Flickr benchmarks, GeoCLIP gains the most, rising from 5.6\% to 17.3\% at 1\,km on Im2GPS3k and from 7.8\% to 30.1\% on YFCC4k. Here many samples land close to the true location, and the base decoder's kilometer-scale error decides whether they fall within 1\,km. The first two steps give most of the gain (10.1 of 11.7 points on Im2GPS3k and 19.8 of 22.3 on YFCC4k). The SatCLIP and TTE decoders are already accurate below 1\,km, so refinement changes their results by at most one point. The step count therefore trades a small cost for accuracy at the finest scale, with diminishing returns after two steps.
\begin{table*}[t]
\centering
\setlength{\tabcolsep}{2.5pt}
\renewcommand{\arraystretch}{1.1}
\footnotesize
\caption{\textbf{Analytic refinement on geolocalization.} XL models trained on each benchmark's training set (IM2GPS3K on MP16-Pro). Every row decodes the same generated samples. We bold improvements over vanilla $> 0.5$.}
\label{tab:refine_benchmarks}
\resizebox{\textwidth}{!}{%
\begin{tabular}{@{}l ccccc|ccccc|ccccc@{}}
\toprule
& \multicolumn{5}{c|}{GeoCLIP} & \multicolumn{5}{c|}{SatCLIP} & \multicolumn{5}{c}{TTE} \\
\cmidrule(lr){2-6} \cmidrule(lr){7-11} \cmidrule(l){12-16}
Decoder & 1\,km & 25\,km & 200\,km & 750\,km & 2500\,km
        & 1\,km & 25\,km & 200\,km & 750\,km & 2500\,km
        & 1\,km & 25\,km & 200\,km & 750\,km & 2500\,km \\
\midrule
\multicolumn{16}{c}{\textit{OSV-5M/OSV-5M}} \\
\midrule
vanilla    & 0.21 & 33.3 & 69.4 & 83.2 & 91.9 & 0.13 & 30.0 & 72.8 & 84.2 & 91.2 & 0.10 & 22.5 & 71.7 & 84.9 & 92.1 \\
\; +1      & 0.26 & 33.4 & 69.4 & 83.2 & 91.9 & 0.12 & 30.3 & 72.8 & 84.2 & 91.2 & 0.09 & 22.7 & 71.8 & 84.9 & 92.1 \\
\; +2      & 0.27 & 33.4 & 69.4 & 83.2 & 91.9 & 0.12 & 30.3 & 72.8 & 84.2 & 91.2 & 0.09 & 22.7 & 71.8 & 84.9 & 92.1 \\
\; +3      & 0.28 & 33.4 & 69.4 & 83.2 & 91.9 & 0.12 & 30.3 & 72.8 & 84.2 & 91.2 & 0.09 & 22.7 & 71.8 & 84.9 & 92.1 \\
\; +4      & 0.27 & 33.4 & 69.4 & 83.2 & 91.9 & 0.12 & 30.3 & 72.8 & 84.2 & 91.2 & 0.09 & 22.7 & 71.8 & 84.9 & 92.1 \\
\; +5      & 0.27 & 33.4 & 69.4 & 83.2 & 91.9 & 0.12 & 30.3 & 72.8 & 84.2 & 91.2 & 0.09 & 22.7 & 71.8 & 84.9 & 92.1 \\
\midrule
\multicolumn{16}{c}{\textit{MP16-Pro/IM2GPS3K}} \\
\midrule
vanilla    & 5.6 & 41.6 & 56.2 & 72.1 & 86.4 & 2.7 & 36.3 & 55.5 & 72.0 & 84.8 & 2.9 & 34.4 & 56.3 & 73.0 & 85.5 \\
\; +1      & \textbf{12.3} & 41.6 & 56.2 & 72.0 & 86.4 & 2.4 & 36.4 & 55.5 & 72.0 & 84.8 & \textbf{3.9} & 34.5 & 56.3 & 73.0 & 85.5 \\
\; +2      & \textbf{15.7} & 41.7 & 56.2 & 72.0 & 86.3 & 2.4 & 36.4 & 55.5 & 72.0 & 84.8 & \textbf{3.9} & 34.5 & 56.3 & 73.0 & 85.5 \\
\; +3      & \textbf{16.8} & 41.7 & 56.2 & 72.0 & 86.3 & 2.4 & 36.4 & 55.5 & 72.0 & 84.8 & \textbf{3.9} & 34.5 & 56.3 & 73.0 & 85.5 \\
\; +4      & \textbf{17.2} & 41.6 & 56.2 & 72.0 & 86.3 & 2.4 & 36.4 & 55.5 & 72.0 & 84.8 & \textbf{3.9} & 34.5 & 56.3 & 73.0 & 85.5 \\
\; +5      & \textbf{17.3} & 41.6 & 56.2 & 72.0 & 86.3 & 2.4 & 36.4 & 55.5 & 72.0 & 84.8 & \textbf{3.9} & 34.5 & 56.3 & 73.0 & 85.5 \\
\midrule
\multicolumn{16}{c}{\textit{YFCC/YFCC4K}} \\
\midrule
vanilla    & 7.8 & 47.4 & 58.3 & 71.1 & 82.9 & 2.7 & 36.5 & 58.4 & 72.0 & 82.7 & 0.9 & 30.0 & 55.3 & 70.8 & 82.7 \\
\; +1      & \textbf{22.6} & 47.4 & 58.3 & 71.1 & 82.9 & 2.7 & 36.6 & 58.4 & 72.0 & 82.7 & 0.9 & 30.1 & 55.3 & 70.8 & 82.7 \\
\; +2      & \textbf{27.6} & 47.5 & 58.3 & 71.1 & 82.9 & 2.7 & 36.6 & 58.4 & 72.0 & 82.7 & 0.9 & 30.1 & 55.3 & 70.8 & 82.7 \\
\; +3      & \textbf{29.1} & 47.5 & 58.3 & 71.1 & 82.9 & 2.7 & 36.6 & 58.4 & 72.0 & 82.7 & 0.9 & 30.1 & 55.3 & 70.8 & 82.7 \\
\; +4      & \textbf{29.9} & 47.5 & 58.3 & 71.1 & 82.9 & 2.7 & 36.6 & 58.4 & 72.0 & 82.7 & 0.9 & 30.1 & 55.3 & 70.8 & 82.7 \\
\; +5      & \textbf{30.1} & 47.5 & 58.3 & 71.1 & 82.9 & 2.7 & 36.6 & 58.4 & 72.0 & 82.7 & 0.8 & 30.1 & 55.3 & 70.8 & 82.7 \\
\bottomrule
\end{tabular}%
}
\end{table*}

\begin{table}[t]
\centering
\caption{\textbf{Analytic refinement across location encoders.} Fraction of 1M uniform land coordinates localized within 1\,km, and decoder cost, as a function of LM refinement steps. The {\setlength{\fboxsep}{1pt}\colorbox{cyan!10}{highlighted}} row is our chosen configuration.}
\small
\setlength{\tabcolsep}{4pt}
\begin{tabular}{@{}lcccccc@{}}
\toprule
& \multicolumn{2}{c}{GeoCLIP} & \multicolumn{2}{c}{SatCLIP} & \multicolumn{2}{c}{TTE} \\
\cmidrule(lr){2-3} \cmidrule(lr){4-5} \cmidrule(lr){6-7}
Decoder & $<$1\,km & GFLOPs & $<$1\,km & GFLOPs & $<$1\,km & GFLOPs \\
\midrule
Base     & 67.91 & 0.076 & 99.65 & 0.076 & 99.59 & 0.076 \\
\; +1    & 99.10 & 0.208 & \hl 100.00 & \hl 0.093 & \hl 100.00 & \hl 0.124 \\
\; +2    & 99.67 & 0.341 & 100.00 & 0.110 & 100.00 & 0.172 \\
\; +3    & 99.79 & 0.473 & 100.00 & 0.127 & 100.00 & 0.220 \\
\; +4    & 99.85 & 0.605 & 100.00 & 0.144 & 100.00 & 0.267 \\
\; +5    & \hl 99.89 & \hl 0.737 & 100.00 & 0.161 & 100.00 & 0.315 \\
\bottomrule
\end{tabular}
\label{tab:lae_refine_all}
\end{table}

\paragraph{Comparing to retrieval.}
Since \name{} generates location embeddings, an alternative to decoding is to retrieve the
location whose embedding is most similar among a gallery covering the Earth. We follow
GeoCLIP~\citep{vivanco2023geoclip}, whose gallery is 100k coordinates sampled from the
training set, and also test a gallery of 100k coordinates sampled uniformly over land.
Table~\ref{tab:decoder_vs_gallery} decodes the same generated embeddings all three ways.
Decoding is best on every benchmark, and the gap is largest on OSV-5M, where the test
locations are kept at a distance from every training location.
\begin{table*}[t]
\centering
\setlength{\tabcolsep}{3pt}
\renewcommand{\arraystretch}{1.1}
\footnotesize
\caption{\textbf{Decoding versus gallery retrieval.} We compare reading out generated embeddings using our LAE decoders with retrieving over training-set or uniform-land location galleries. Reported metrics use \name-XL GeoCLIP model (OSV-5M uses the OSV-5M model; YFCC4k the YFCC-44M model; IM2GPS3K the MP16-Pro model) Best in \textbf{bold}.}
\label{tab:decoder_vs_gallery}
\resizebox{\textwidth}{!}{%
\begin{tabular}{@{}l ccccc|ccccc|ccccc@{}}
\toprule
& \multicolumn{5}{c|}{OSV-5M} & \multicolumn{5}{c|}{YFCC4k} & \multicolumn{5}{c}{IM2GPS3K} \\
\cmidrule(lr){2-6} \cmidrule(lr){7-11} \cmidrule(l){12-16}
Readout & GeoS. & Mean & Ctry & Reg. & City & 1\,km & 25\,km & 200\,km & 750\,km & 2500\,km & 1\,km & 25\,km & 200\,km & 750\,km & 2500\,km \\
\midrule
Decoder                      & \textbf{4095} & \textbf{852} & \textbf{83.4} & \textbf{60.1} & \textbf{24.6} & \textbf{30.1} & \textbf{47.5} & \textbf{58.3} & \textbf{71.1} & \textbf{82.9} & \textbf{17.3} & \textbf{41.6} & \textbf{56.2} & \textbf{72.0} & \textbf{86.3} \\
Training set gallery  & 4040 & 878 & 82.2 & 56.2 & 19.9 & 28.4 & 46.7 & 58.0 & 71.0 & \textbf{82.9} & \textbf{17.3} & 41.4 & \textbf{56.2} & 71.7 & \textbf{86.3} \\
Uniform-land gallery         & 4067 & 861 & 82.3 & 55.9 & 17.5 & 0.0 & 22.8 & 55.4 & 70.1 & 82.5 & 0.0 & 17.4 & 54.0 & 71.0 & 85.9 \\
\bottomrule
\end{tabular}%
}
\end{table*}

\subsection{\name Additional Details}
\label{app:ldm}

\subsubsection{\name Implementation Details}
\label{app:ldm:implementation}
\begin{table}[t]
  \centering
  \small
  \caption{Diffusion model sizes. Widths and depths follow DiT~\citep{peebles2023dit};
  the MLP hidden size is $4w$. Parameter counts are for a $512$-dimensional location  embedding and a $2048$-dimensional image condition, and exclude the REPA head ($2$--$3$M parameters).}
  \label{tab:sizes}
  \begin{tabular}{lccccc}
    \toprule
    Model & Width $w$ & Depth $D$ & MLP hidden & Params \\
    \midrule
    B  & 768  & 12 & 3072 & 82M    \\
    L  & 1024 & 24 & 4096 & 284M  \\
    XL & 1152 & 28 & 4608 & 417M \\
    \bottomrule
  \end{tabular}
\end{table}

This section gives the details for \name{} as trained for the main results. Notation follows Sec.~\ref{sec:ldm}.

\paragraph{Latent standardization.}
The diffusion operates on a standardized embedding: with $\mu, \sigma \in \mathbb{R}^d$ the
per-dimension mean and standard deviation of $E(\ell)$ over the training coordinates, the model
sees $x = (E(\ell) - \mu)/\sigma$, and samples are de-standardised before decoding. This matters
for GeoCLIP, whose raw embeddings have per-dimension standard deviation ${\approx}0.014$, two
orders of magnitude below the unit-variance noise the path assumes.

\paragraph{Time sampling.}
Training times are not uniform. We draw $u \sim \mathcal{U}(0,1)$ and warp it through the
sigmoid schedule of \citet{dufour2025around},
\begin{equation}
  t(u) = \frac{\sigma(b) - \sigma\big(a + (b-a)\,u\big)}{\sigma(b) - \sigma(a)},
  \qquad a = -5,\; b = 5,
  \label{eq:schedule}
\end{equation}
which places more mass near both ends of the path ($t \approx 0$ and $t \approx 1$) than in the
middle. The same warp defines the sampling grid: the $N$ Euler steps are equally spaced in $u$,
hence non-uniform in $t$, with $t$ clamped to $\le 1 - 10^{-3}$ so the velocity conversion of
Sec.~\ref{sec:ldm} stays finite. \citet{dufour2025around} use the asymmetric setting
$(a,b) = (-7, 3)$; the symmetric one above performed better in our schedule ablation.

\paragraph{Representation alignment.}
The similarity in $\mathcal{L}_{\text{REPA}}$ has a value term and a direction term,
\begin{equation}
  \mathrm{sim}(a, x) = -\,\|a - x\|_2^2 \;+\; \lambda_{\cos}\,\frac{a^{\top} x}{\|a\|\,\|x\|},
  \qquad \lambda_{\cos} = 0.5,
  \label{eq:sim}
\end{equation}
so that $\mathcal{L}_{\text{REPA}} = \mathbb{E}\big[\|h_\phi(f_k) - x\|^2 + \lambda_{\cos}(1 -
\cos(h_\phi(f_k), x))\big]$ up to a constant. REPA as published uses the cosine alone; we add
the squared error so that $h_\phi(f_k)$ matches $x$ in magnitude as well as direction, i.e.\ it
is an $x$-prediction at depth $k$. The head reads block $k = 8$ and is
$\mathrm{LN} \to \mathrm{Linear}(w \to 2048) \to \mathrm{SiLU} \to \mathrm{Linear}(2048 \to d)$.
We use $\lambda = 0.5$ and discard the head at inference.

\paragraph{Conditioning vector.}
A 288-channel sinusoidal embedding of the timestep (two-layer SiLU MLP to width $w$) is
summed with a linear projection of the image features $c \in \mathbb{R}^{2048}$
(StreetCLIP $\Vert$ DINOv3 CLS tokens, precomputed); the null condition $\varnothing$ is
the projection's bias. Each block maps the sum to AdaLN scale, shift and gate via
$\mathrm{SiLU} \to \mathrm{Linear}(w \to 3w)$.

\begin{table}[h]
\centering\small
\caption{Training configuration of the XL model.}
\label{tab:app_train}
\begin{tabular}{ll}
\toprule
Latent $d$ & 512 (GeoCLIP), standardized per dimension \\
Width $w$ / blocks / MLP expansion & 1152 / 28 / 4 \\
Parameters & 420M \\
Alignment & $k = 8$, $\lambda = 0.5$, $\lambda_{\cos} = 0.5$ \\
Time schedule & Eq.~\ref{eq:schedule}, $(a, b) = (-5, 5)$ \\
Condition dropout $p_{\text{drop}}$ & 0.1 \\
Optimizer & AdamW, $\beta = (0.9, 0.999)$, weight decay 0.05 \\
Learning rate & $8\times10^{-4}$; linear warm-up 1k steps; cosine decay to 0 \\
Batch size / steps & 1024 / 1M \\
Gradient clipping & global norm 1.0 \\
Precision & bf16 autocast \\
EMA & decay 0.9999; used for all evaluation \\
\bottomrule
\end{tabular}
\end{table}

\subsubsection{Generative Comparison.}

Table~\ref{tab:generative_metrics} evaluates unconditional generation using the precision, recall, density, and coverage metrics from PLONK. On OSV-5M, \name achieves the highest precision and density but trails the coordinate-space baselines in recall and coverage. A random sample of OSV-5M training locations obtains 0.950 precision, 0.791 recall, 0.923 density, and 0.385 coverage, closely matching all three \name variants. This suggests that the lower recall and coverage arise from the shift between the training and test distributions rather than mode dropping. For example, the United States accounts for 24.9\% of the training set but only 11.4\% of the test set, and the country distributions have a total variation distance of 0.46. Matching the training distribution therefore undersamples regions overrepresented in the test set. The baselines trade precision and density for higher recall and coverage by producing a broader distribution. On YFCC4k, the metrics saturate, and all methods are within noise of the ground-truth ceiling.
\begin{table*}[t]
\centering
\setlength{\tabcolsep}{6pt}
\renewcommand{\arraystretch}{1.1}
\scriptsize
\caption{\textbf{Generative metrics of the unconditional distribution.} Precision, recall,
density and coverage~\cite{naeem2020reliable} of samples drawn from $p(\mathbf{y}\,|\,\varnothing)$
against the test-set locations of OSV-5M (210k) and YFCC4k (4.5k), following \cite{dufour2025around}. Best in \textbf{bold}, second \textit{italicized}.}
\label{tab:generative_metrics}
\begin{tabular}{@{}l cccc cccc@{}}
\toprule
& \multicolumn{4}{c}{OSV-5M} & \multicolumn{4}{c}{YFCC4k} \\
\cmidrule(lr){2-5}\cmidrule(l){6-9}
Method & precision$\uparrow$ & recall$\uparrow$ & density$\uparrow$ & coverage$\uparrow$
       & precision$\uparrow$ & recall$\uparrow$ & density$\uparrow$ & coverage$\uparrow$ \\
\midrule
Uniform                       & 0.29  & 0.98  & 0.21  & 0.21
                              & 0.59  & 0.99  & 0.38  & 0.22  \\
vMF                           & 0.598 & \textbf{0.982} & 0.499 & 0.446
                              & 0.667 & \textbf{0.993} & 0.542 & 0.599 \\
vMFMix                        & 0.513 & \textit{0.980} & 0.422 & 0.358
                              & 0.626 & \textit{0.988} & 0.474 & 0.498 \\
Diff $\mathbb{R}^3$           & 0.822 & 0.916 & 0.752 & 0.568
                              & 0.938 & 0.959 & 0.959 & 0.837 \\
FM $\mathbb{R}^3$             & 0.845 & 0.907 & 0.799 & \textit{0.575}
                              & \textit{0.953} & 0.959 & 1.037 & \textit{0.920} \\
RFM $\mathcal{S}_2$           & 0.841 & 0.896 & 0.797 & \textbf{0.590}
                              & \textbf{0.957} & 0.952 & \textit{1.060} & \textbf{0.926} \\
LocDiff                       & -- & -- & 0.795 & 0.572
                              & -- & -- & \textbf{1.072} & 0.915 \\
\midrule
\textbf{Ours} (GeoCLIP)       & \textbf{0.953} & 0.792 & \textbf{0.932} & 0.351
                              & 0.946 & 0.934 & 1.032 & 0.879 \\
\textbf{Ours} (SatCLIP)       & \textit{0.951} & 0.779 & \textit{0.925} & 0.378
                              & 0.946 & 0.930 & 1.021 & 0.879 \\
\textbf{Ours} (TTE)           & 0.950 & 0.794 & 0.920 & 0.383
                              & 0.945 & 0.932 & 1.019 & 0.875 \\
\bottomrule
\end{tabular}
\end{table*}

\subsubsection{Leveraging the Posterior for Prediction}
\label{app:posterior}

Every prediction in the main tables is a single draw from the \name's conditional distribution. A single draw can land in a minor mode or in the tail of the correct one, and it says nothing about how certain the model is. Drawing several samples exposes that structure. Where the samples agree the model is confident, and where they split between regions the split itself is informative. Here we draw 16 samples per test image with
different seeds and test rules for combining them into one prediction.

\paragraph{Aggregation rules.}
\emph{Single sample} uses the first draw and matches the main-paper protocol.
\emph{Center} is the spherical mean of the 16 decoded coordinates. The remaining rules
first cluster the samples, joining any two within 200\,km of each other, and keep only the
cluster with the most samples. \emph{Cluster mean} is the mean coordinate of that cluster.
\emph{Cluster medoid} is the sample closest to the cluster mean. \emph{Cluster latent
mean} averages the generated location embeddings of that cluster and decodes the result.

\paragraph{Results.}
Table~\ref{tab:aggregation} shows the same pattern for all three encoders. Center lowers
the mean error on OSV-5M but hurts accuracy at nearly every threshold, since averaging
across modes pulls the prediction between them. Restricting to the largest cluster fixes
this, and the three cluster rules improve on the single sample on OSV-5M and YFCC4k at
almost every metric, with smaller and less consistent gains on Im2GPS3k. Cluster mean is
best at 200\,km for every encoder, gaining 2.6 to 3.6 points on YFCC4k. Cluster medoid is
best at 25\,km for every encoder, raising GeoCLIP from 47.5 to 49.5 on YFCC4k and from
41.6 to 42.8 on Im2GPS3k, and from 30.1 to 30.8 at 1\,km on YFCC4k. On OSV-5M the cluster
rules add 23 to 47 GeoScore and 0.5 to 1.1 points of country accuracy. Sixteen samples
therefore give a consistent improvement at 16 times the sampling cost, and cluster medoid,
the rule that returns an actual sample, is the safest choice.

\begin{table*}[t]
\centering
\setlength{\tabcolsep}{3pt}
\renewcommand{\arraystretch}{1.1}
\footnotesize
\caption{\textbf{Aggregating 16 posterior samples.} Each test image is sampled 16 times with different seeds, and predictions are combined by the rule in each row. OSV-5M uses the OSV-5M models; YFCC4k the YFCC-44M models; IM2GPS3K the MP16-Pro models. Best rule per encoder in \textbf{bold}.}
\label{tab:aggregation}
\resizebox{\textwidth}{!}{%
\begin{tabular}{@{}l ccccc|ccccc|ccccc@{}}
\toprule
& \multicolumn{5}{c|}{OSV-5M} & \multicolumn{5}{c|}{YFCC4k} & \multicolumn{5}{c}{IM2GPS3K} \\
\cmidrule(lr){2-6} \cmidrule(lr){7-11} \cmidrule(l){12-16}
Rule & GeoS. & Mean & Ctry & Reg. & City & 1\,km & 25\,km & 200\,km & 750\,km & 2500\,km & 1\,km & 25\,km & 200\,km & 750\,km & 2500\,km \\
\midrule
\multicolumn{16}{@{}l}{\textit{GeoCLIP}} \\
Single sample        & 4095 & 851 & 83.4 & 60.1 & 24.6 & 30.1 & 47.5 & 58.3 & 71.1 & 82.9 & 17.3 & 41.6 & 56.2 & 72.0 & 86.3 \\
Center               & 4081 & \textbf{767} & 82.6 & 58.7 & 24.2 & 26.1 & 41.5 & 54.8 & 69.3 & 82.1 & 16.4 & 38.5 & 55.1 & 72.2 & 86.3 \\
Cluster mean         & \textbf{4142} & 790 & \textbf{84.5} & 61.1 & 25.4 & 28.5 & 48.5 & \textbf{61.9} & \textbf{72.6} & \textbf{83.3} & 17.5 & 42.0 & \textbf{58.1} & \textbf{73.3} & \textbf{86.7} \\
Cluster latent mean  & 4141 & 790 & \textbf{84.5} & \textbf{61.2} & \textbf{25.5} & 30.3 & 49.0 & 61.8 & \textbf{72.6} & \textbf{83.3} & 17.9 & 42.5 & \textbf{58.1} & \textbf{73.3} & \textbf{86.7} \\
Cluster medoid       & 4141 & 790 & 84.4 & 61.1 & 25.3 & \textbf{30.8} & \textbf{49.5} & 61.6 & \textbf{72.6} & \textbf{83.3} & \textbf{18.1} & \textbf{42.8} & 58.0 & 73.2 & \textbf{86.7} \\
\midrule
\multicolumn{16}{@{}l}{\textit{SatCLIP}} \\
Single sample        & 4125 & 912 & 83.4 & 61.0 & 22.4 & 2.7 & 36.6 & 58.4 & 72.0 & 82.7 & 2.4 & 36.4 & 55.5 & \textbf{72.0} & \textbf{84.8} \\
Center               & 4103 & \textbf{848} & 82.7 & 60.2 & 22.3 & 2.5 & 32.7 & 55.3 & 70.5 & 83.1 & \textbf{3.1} & 35.9 & 55.0 & 71.0 & 84.7 \\
Cluster mean         & \textbf{4148} & 860 & \textbf{83.9} & \textbf{61.4} & 22.8 & \textbf{2.8} & 37.3 & \textbf{61.0} & \textbf{73.2} & \textbf{84.0} & \textbf{3.1} & \textbf{36.6} & \textbf{55.8} & 71.5 & 84.7 \\
Cluster latent mean  & \textbf{4148} & 860 & \textbf{83.9} & \textbf{61.4} & \textbf{22.9} & \textbf{2.8} & 37.2 & 60.9 & \textbf{73.2} & \textbf{84.0} & \textbf{3.1} & \textbf{36.6} & \textbf{55.8} & 71.6 & 84.7 \\
Cluster medoid       & \textbf{4148} & 860 & \textbf{83.9} & \textbf{61.4} & 22.8 & 2.6 & \textbf{37.6} & 60.7 & 73.1 & \textbf{84.0} & 2.6 & \textbf{36.6} & 55.7 & 71.5 & 84.7 \\
\midrule
\multicolumn{16}{@{}l}{\textit{TTE}} \\
Single sample        & 4129 & 849 & 83.5 & 57.4 & 18.4 & 0.9 & 30.1 & 55.3 & 70.8 & 82.7 & \textbf{3.9} & 34.5 & 56.3 & 73.0 & \textbf{85.5} \\
Center               & 4112 & \textbf{783} & 82.9 & 56.8 & 18.2 & 0.9 & 27.4 & 52.8 & 70.0 & 82.5 & 3.6 & 32.8 & 54.8 & 72.3 & 84.7 \\
Cluster mean         & \textbf{4157} & 794 & \textbf{84.1} & \textbf{58.1} & \textbf{18.7} & 1.0 & 30.9 & \textbf{58.8} & 72.7 & \textbf{83.5} & 3.8 & 33.9 & 56.6 & 73.0 & 85.3 \\
Cluster latent mean  & \textbf{4157} & 794 & \textbf{84.1} & \textbf{58.1} & \textbf{18.7} & 1.0 & 31.0 & 58.7 & 72.8 & \textbf{83.5} & 3.8 & 34.3 & 56.5 & 73.0 & 85.3 \\
Cluster medoid       & 4156 & 794 & \textbf{84.1} & \textbf{58.1} & \textbf{18.7} & \textbf{1.1} & \textbf{31.6} & 58.4 & \textbf{72.9} & 83.4 & 3.8 & \textbf{34.8} & \textbf{56.7} & \textbf{73.1} & 85.4 \\
\bottomrule
\end{tabular}%
}
\end{table*}

\subsubsection{\name with GeoRanker}
\label{app:georanker}

\begin{table}[t]
\centering
\setlength{\tabcolsep}{2.5pt}
\renewcommand{\arraystretch}{1.0}
\scriptsize
\caption{\textbf{\name{} as a prior for GeoRanker.} Accuracy (\%) within distance
thresholds and median error (km). GeoRanker and \name-XL (GeoCLIP) are trained on MP16-Pro.}
\label{tab:georanker}
\begin{tabular}{lcccccr@{\hspace{10pt}}cccccr}
\toprule
& \multicolumn{6}{c}{\textbf{Im2GPS3k}} & \multicolumn{6}{c}{\textbf{YFCC4k}} \\
\cmidrule(lr){2-7}\cmidrule(lr){8-13}
Method & 1\,km & 25\,km & 200\,km & 750\,km & 2500\,km & Med. & 1\,km & 25\,km & 200\,km & 750\,km & 2500\,km & Med. \\
\midrule
GeoRanker                      & 18.6 & 45.0 & 61.5 & 76.2 & 89.2 & 63.5 & 32.6 & 43.3 & 53.8 & 70.0 & 82.6 & 121.2 \\
\midrule
GeoRanker + prior              & 18.9 & 46.5 & 62.9 & 76.9 & 89.4 & 47.7 & 33.4 & 45.4 & 55.8 & 71.4 & 83.4 &  79.9 \\
GeoRanker + prior + samples    & 19.6 & 47.4 & 63.5 & 77.9 & \textbf{89.6} & 38.0 & 33.6 & \textbf{45.5} & 56.4 & \textbf{72.2} & \textbf{84.1} &  75.4 \\
GeoRanker + prior + samples + snap & \textbf{20.1} & \textbf{47.6} & \textbf{64.1} & \textbf{78.1} & \textbf{89.6} & \textbf{36.0} & \textbf{34.0} & \textbf{45.5} & \textbf{56.8} & 71.9 & 83.8 & \textbf{73.7} \\
\bottomrule
\end{tabular}
\end{table}

\paragraph{GeoRanker.}
GeoRanker~\citep{jia2025georanker} is a learned reranker over a small set of candidate locations. For each query image, the candidates are retrieved MP16-Pro~\citep{jia2024g3} images with their coordinates and place names, plus GPS guesses from GPT-4V~\citep{openai2024gpt4}. A fine-tuned Qwen2-VL-7B~\citep{wang2024qwen2} scores each candidate against the query with a
scalar reward trained to be high for candidates near the true location, and the highest-reward candidate is used for coordinate prediction. We use the released weights, prompts and candidate lists (12 retrievals and 3 guesses) without modification and reproduce their reported accuracies in Table~\ref{tab:georanker}.

\paragraph{Injecting \name{}.}
We draw $K = 16$ samples $\hat{\ell}_1, \dots, \hat{\ell}_K$ per image from \name-XL (GeoCLIP) trained on MP16-Pro and use them both as candidates and as a prior over the candidates. As candidates, each sample is added to
the pool as a GPS-only guess and scored by the frozen reranker. As a prior, we add to every candidate's reward the fraction of samples within $R$\,km of it,
\begin{equation}
  s(c) = r(q, c) + \lambda \, \frac{1}{K} \sum_{k=1}^{K}
  \mathbb{1}\big[\, d_{\mathcal{S}^2}(c, \hat{\ell}_k) \le R \,\big],
  \qquad \hat{\ell} = \arg\max_{c \in \mathcal{C}} s(c),
  \label{eq:fusion}
\end{equation}
where $\mathcal{C}$ is the candidate pool and $d_{\mathcal{S}^2}$ the great-circle distance. We use $\lambda = 1$ and $R = 50$\,km. We also try snapping each sample to its nearest MP16-Pro training point and add that image with its place name as a retrieval-style
candidate. We find that snapping improves performance, likely because of the high geographic overlap between the training and test sets (Table~\ref{tab:train_test_overlap}). GeoRanker is never
fine-tuned, and since rewards are pointwise every candidate is scored once, so all rows of Table~\ref{tab:georanker} share one set of rewards.

\paragraph{Results.}
Table~\ref{tab:georanker} reports Im2GPS3k and YFCC4k. 
Adding \name{} as a
prior improves GeoRanker at every threshold on both benchmarks and cuts the median error
from 63.5 to 47.7\,km and from 121.2 to 79.9\,km. Adding the samples as candidates
improves it further, and snapping adds a small gain at fine thresholds. The full system
gains 2.6 and 2.2 points at 25\,km, 2.6 and 3.0 points at 200\,km, and reduces the median
error by 43\% and 39\%.

\subsection{Datasets}
\label{app:datasets}

\subsubsection{Metrics}

\paragraph{Distance error.}
All metrics start from the great-circle distance $d$ in km between the predicted and the
ground-truth coordinate on a sphere of radius 6371\,km. We report the mean and the median
of $d$ over the test set. The mean is dominated by the few predictions on the wrong
continent, and the median by the typical case.

\paragraph{GeoScore.}
Following OSV-5M~\citep{astruc2024openstreetview}, each prediction receives
$5000 \cdot\exp(-d / 1492.7)$ and GeoScore is the average over the test set. It ranges from 0 to
5000 and rewards fine-grained accuracy while remaining bounded for large errors.

\paragraph{Accuracy within a radius.}
Accuracy at $r$\,km is the percentage of test images with $d < r$. On Im2GPS3k and YFCC4k we
use the standard radii of 1, 25, 200, 750 and 2500\,km~\citep{hays2008im2gps}, corresponding
to street, city, region, country and continent level.

\paragraph{Administrative accuracy.}
On OSV-5M we follow \citet{dufour2025around} and reverse-geocode the predicted coordinate offline to a country, region and city with GeoNames, and count a prediction as
correct when its label matches the ground-truth label at that level.

\subsubsection{Dataset Distributions}
\label{app:dataset_comparison}
\begin{figure}[H]
    \centering
    \includegraphics[width=\textwidth]{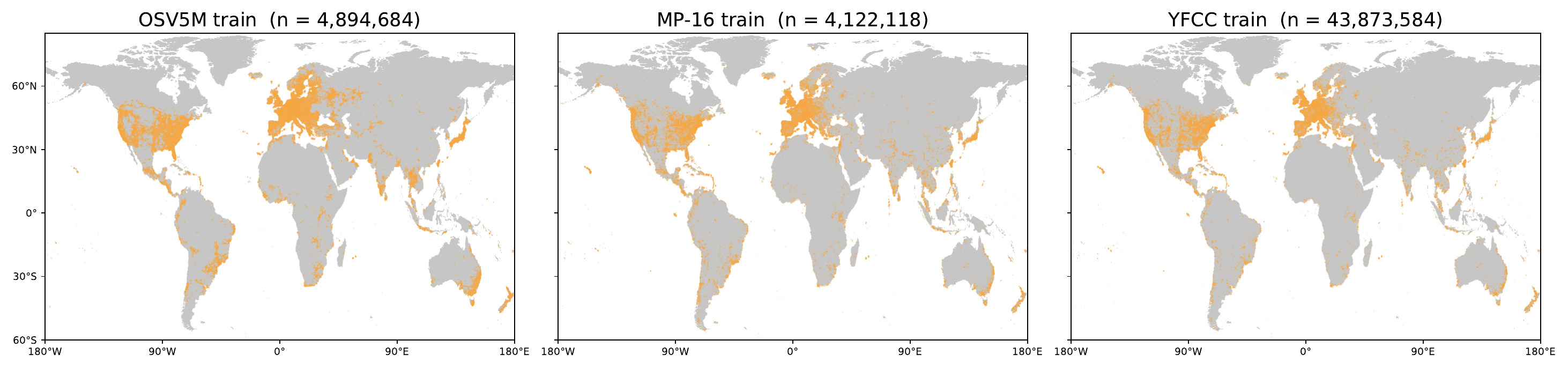}
    \caption{Geographic distribution of training sets.}
    \label{fig:train_dist}
\end{figure}

\begin{figure}[H]
    \centering
    \includegraphics[width=\textwidth]{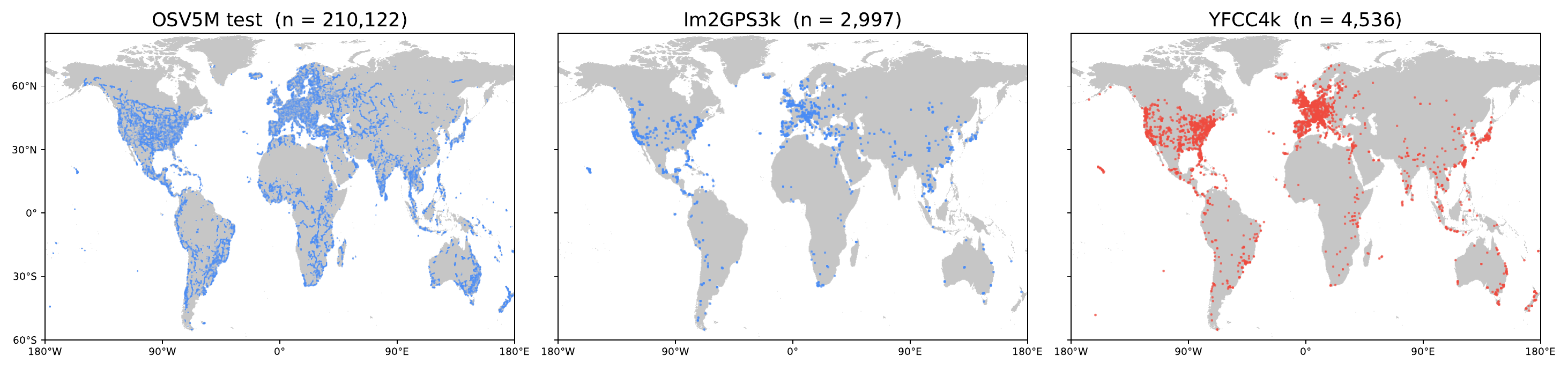}
    \caption{Geographic distribution of test sets.}
    \label{fig:test_dist}
\end{figure}

\begin{table}[t]
\centering
\setlength{\tabcolsep}{4pt}
\renewcommand{\arraystretch}{1.1}
\small
\caption{\textbf{Train--test geographic overlap.} For every test location, the great-circle distance to its nearest training location. ``Within'' columns give the share of test locations with a training location inside that radius.}
\label{tab:train_test_overlap}
\begin{tabular}{@{}lcccc@{}}
\toprule
& \multicolumn{2}{c}{Distance (km)} & \multicolumn{2}{c}{Within (\%)} \\
\cmidrule(lr){2-3} \cmidrule(l){4-5}
Training set & Median & Mean & 1\,km & 25\,km \\
\midrule
\rowcolor{gray!20}\multicolumn{5}{@{}l}{\textit{Test: OSV-5M}} \\
OSV-5M       & 4.32  & 15.7 & 0.1  & 86.4 \\
\midrule
\rowcolor{gray!20}\multicolumn{5}{@{}l}{\textit{Test: Im2GPS3k}} \\
MP16-Pro     & 0.081 & 1.03 & 86.5 & 99.0 \\
YFCC-44M     & 0.022 & 0.82 & 92.0 & 99.1 \\
\midrule
\rowcolor{gray!20}\multicolumn{5}{@{}l}{\textit{Test: YFCC4k}} \\
MP16-Pro & 0.004 & 0.69 & 93.7 & 99.7 \\
YFCC-44M     & 0.000 & 0.19 & 97.2 & 99.8 \\
\bottomrule
\end{tabular}
\end{table}

\paragraph{Geographic bias.}
Figure~\ref{fig:train_dist} shows the geographic distribution of each training set. North America and Europe account for 70\% of OSV-5M, 77\% of MP16-Pro, and 81\% of YFCC-44M. Africa accounts for 4.0\% of OSV-5M and less than 2\% of either Flickr dataset. We assign continents using Natural Earth admin-0 polygons. The test distributions differ substantially (Figure~\ref{fig:test_dist}). YFCC4k closely follows its training distribution, with 77\% of locations in North America and Europe and 1.7\% in Africa. Im2GPS3k remains concentrated in North America and Europe at 61\%. The OSV-5M test set is more balanced across countries. North America and Europe account for 39\%, while Africa and South America account for 17\% and 12\%, respectively. OSV-5M therefore evaluates a larger geographic distribution shift, while the Flickr benchmarks remain closer to their training distributions.

\paragraph{Train--test spatial overlap.}
Table~\ref{tab:train_test_overlap} reports the distance from each test location to its nearest training location. The Flickr benchmarks have substantial spatial overlap with their training sets. The median distance is 81\,m for Im2GPS3k/MP16-Pro and tens of metres for YFCC4k, with 87--97\% of test locations falling within 1\,km of a training location. This overlap follows from their shared Flickr provenance, where photos often cluster around the same landmarks and viewpoints. Retrieval from a training gallery and classification over cells anchored at training coordinates benefit directly from this overlap. Sub-kilometre accuracy on these benchmarks therefore depends in part on training-set coverage. OSV-5M has much less overlap. Only 0.1\% of its test locations fall within 1\,km of a training location, and the median nearest-training distance is 4.3\,km. Copying training coordinates is therefore insufficient for street-level accuracy. Together with the geographic rebalancing described above, this difference helps explain the stronger relative performance of retrieval methods on Im2GPS3k and YFCC4k.
\begin{table}[t]
\centering
\setlength{\tabcolsep}{5pt}
\renewcommand{\arraystretch}{1.0}
\scriptsize
\caption{Reported training sources of baselines on IM2GPS3k and YFCC4k.}
\label{tab:training_sources}
\begin{tabular}{@{}l cc cc@{}}
\toprule
& \multicolumn{2}{c}{IM2GPS3k} & \multicolumn{2}{c}{YFCC4k} \\
\cmidrule(lr){2-3}\cmidrule(l){4-5}
Method & Source & \# Images & Source & \# Images \\
\midrule
PlaNet~\citep{weyand2016planet}                 & Web & 126M &  Web & 126M \\
CPlaNet~\citep{seo2018cplanet}                  & Flickr & 30.3M & Flickr & 30.3M  \\
Translocator~\citep{pramanick2022translocator}  & MP-16 & 4.7M & MP-16 & 4.7M \\
GeoDecoder~\citep{clark2023geodecoder}          & MP-16 & 4.7M & MP-16 & 4.7M \\
PIGEOTTO~\citep{haas2024pigeon}                 & Flickr/Wikipedia & 4M & Flickr/Wikipedia & 4M \\
GeoCLIP~\citep{vivanco2023geoclip}              & MP-16 & 4.7M & MP-16 & 4.7M \\
HierLoc~\citep{gadi2026hierloc}                 & MP-16 & 4.7M & MP-16 & 4.7M \\
Img2Loc~\citep{zhou2024img2loc}                 & training-free & -- & training-free & -- \\
G3~\citep{jia2024g3}                            & MP16-Pro & 4.1M & MP16-Pro & 4.1M \\
GeoRanker~\citep{jia2025georanker}              & MP16-Pro & 4.1M & MP16-Pro & 4.1M \\
\midrule
vMFMix~\citep{izbicki2019exploiting}                                & -- & -- & YFCC & 48M \\
vMF~\citep{dufour2025around}                                        & -- & -- & YFCC & 48M \\
Diff $\mathbb{R}^3$~\citep{dufour2025around}                        & YFCC & 48M & YFCC & 48M \\
FM $\mathbb{R}^3$~\citep{dufour2025around}                          & YFCC & 48M & YFCC & 48M \\
RFM $\mathcal{S}_2$~\citep{dufour2025around}                        & -- & -- & -- & -- \\
RFM\textsubscript{10M} $\mathcal{S}_2$~\citep{dufour2025around}     & YFCC & 48M & YFCC & 48M \\
LocDiff~\citep{wang2025locdiff}                                     & MP-16 & 4.7M & YFCC & 48M \\
LocDiff-H~\citep{wang2025locdiff}                                   & MP-16 & 4.7M & -- & -- \\
\bottomrule
\end{tabular}
\end{table}

\subsubsection{Dataset Filtering}

\paragraph{YFCC-44M.}
We build YFCC-44M from the public YFCC100M metadata dump~\cite{thomee2016yfcc100m}, distributed by Multimedia Commons as a SQLite file, following the filtering of~\cite{dufour2025around,wang2025locdiff}. We keep photos, discarding videos, that have coordinates with a Flickr geotag accuracy of at least 12 on Flickr's 1--16 scale, where 16 is street level. This leaves 43.90M photos. We remove the YFCC4k test set by Flickr photo ID. We download the remaining images from the Multimedia Commons S3 mirror and discard those that are unavailable. Finally, we remove by photo ID the Im2GPS3k test photos that also appear in YFCC100M, leaving 43,873,584 training images.

\begin{wrapfigure}{r}{0.6\textwidth}
    \centering
    \includegraphics[width=\linewidth]{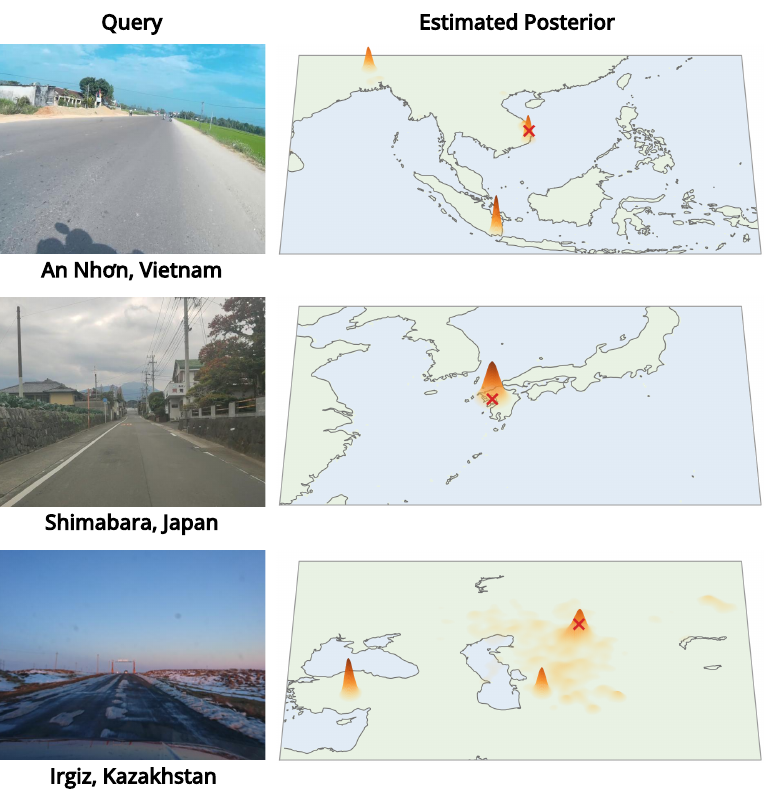}
    \caption{\textbf{Estimated posterior distribution.} We use \name-XL (GeoCLIP) trained on OSV-5M.}
    \label{fig:main_posteriors}
\end{wrapfigure}

\subsection{Additional Qualitative Results}
\label{app:qualitative}

Figure~\ref{fig:main_posteriors} visualizes \name's predictive distribution by decoding
Monte Carlo samples into a density over the globe. The posterior adapts its shape to the
image. For the Vietnamese roadside, it places several modes on visually consistent
tropical regions, including the correct coast; for the Japanese street, whose vegetation
and architecture are distinctive, it collapses to a single sharp mode; for the Kazakh
steppe, it spreads over the plausible region, keeping considerable mass near the ground
truth while also entertaining a lookalike mode in Turkey.

\end{document}